%% file: main.tex
\documentclass[10pt]{article} 
\usepackage[preprint]{neurips_2023}

\input{math_commands.tex}

\usepackage{hyperref}
\usepackage{url}

\usepackage{hyperref}
\usepackage{url}
\usepackage{graphicx}
\usepackage{algorithm}
\usepackage{algorithmic}
\usepackage{caption}
\usepackage{subcaption}
\usepackage{wrapfig}
\usepackage{amsmath}
\usepackage{amssymb}

\title{Adversarial Style Transfer for Robust Policy Optimization in Deep Reinforcement Learning}

\author{%
  Md Masudur Rahman and Yexiang Xue\\
  Department of Computer Science\\
  Purdue University\\
  West Lafayette, IN 47907, USA \\
  \texttt{\{rahman64,yexiang\}@purdue.edu} \\
}

\def\month{MM}  
\def\year{YYYY} 
\def\openreview{\url{https://openreview.net/forum?id=XXXX}} 

\begin{document}

\maketitle

\input{paper}

\bibliography{main}
\bibliographystyle{tmlr}

\appendix
\section*{Appendix}

\input{appendix} 

\end{document}

%% file: math_commands.tex
\usepackage{amsmath,amsfonts,bm}

\def\eqref#1{equation~\ref{#1}}

\def\1{\bm{1}}

\DeclareMathAlphabet{\mathsfit}{\encodingdefault}{\sfdefault}{m}{sl}
\SetMathAlphabet{\mathsfit}{bold}{\encodingdefault}{\sfdefault}{bx}{n}



%% file: paper.tex
\begin{abstract}
This paper proposes an algorithm that aims to improve generalization for reinforcement learning agents by removing overfitting to confounding features. Our approach consists of a max-min game theoretic objective. A generator transfers the style of observation during reinforcement learning. An additional goal of the generator is to perturb the observation, which maximizes the agent's probability of taking a different action. In contrast, a policy network updates its parameters to minimize the effect of such perturbations, thus staying robust while maximizing the expected future reward. Based on this setup, we propose a practical deep reinforcement learning algorithm, Adversarial Robust Policy Optimization (ARPO), to find a robust policy that generalizes to unseen environments. We evaluate our approach on Procgen and Distracting Control Suite for generalization and sample efficiency.
Empirically, ARPO shows improved performance compared to a few baseline algorithms, including data augmentation.
\end{abstract}

\input{intro.tex}

\input{background.tex}
\input{method.tex}

\input{experiment.tex}

\input{relatedwork.tex}

\input{conclusion.tex}

%% file: intro.tex
\section{Introduction}
The reinforcement learning (RL) environments often provide observation, which is a high-dimensional projection of the true state, complicating policy learning as the deep neural network model might mistakenly correlate reward with irrelevant information.  
Thus deep neural networks might overfit irrelevant features in the training data due to their high flexibility and the long-stretched time duration of the RL training process, which leads to poor generalization \citep{hardt2016train,zhang2018dissection,cobbe2019quantifying,cobbe2020leveraging,zhang2018study,machado2018revisiting,gamrian2019transfer}. 
These irrelevant features (e.g., background color) usually do not impact the reward; thus, an optimal agent should avoid focusing on them during policy learning.  Even worse, this might lead to incorrect state representations, which prevent deep RL agents from performing well even in slightly different environments. Thus, to learn a correct state representation, the agent needs to avoid those features.

\begin{figure}[t]
    \centering
    \includegraphics[width=0.7\linewidth]{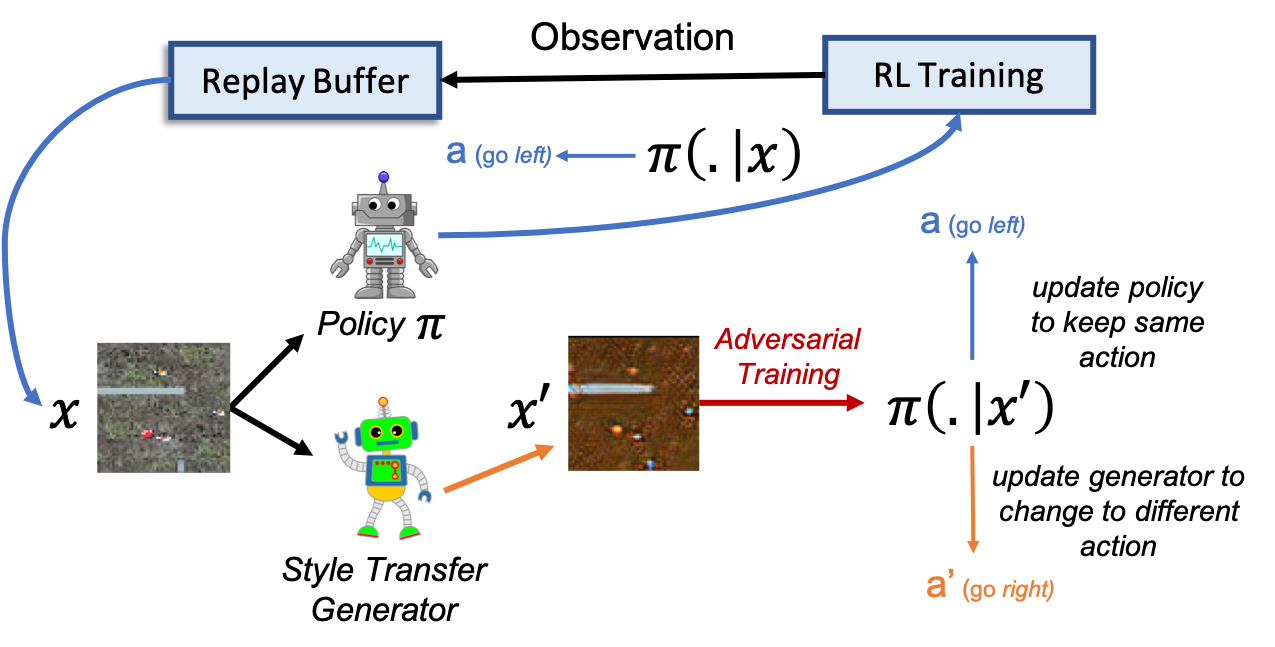}
    \caption{\small  Overview of our approach. The method consists of a minimax game theoretic objective. It first applies a clustering approach to generate $n$ group of clusters based on the different visual features in the observation. A generator is then used to style-translate observation from one cluster to another while maximizing the change in action probabilities of the policy. In contrast, the policy network updates its policy parameters to minimize the translation effect while optimizing for the RL objective that maximizes the expected future reward. 
    }
    \label{fig:framework}
\end{figure}

In recent times, generalization in RL has been explored extensively. In zero-shot generalization framework \citep{zhang2018study,Song2020Observational,wang2019generalization,packer2018assessing} the agent is trained on a finite training set of MDP levels and then evaluated on the full distribution of the MDP levels. 
The distribution of the MDP levels can be generated in various ways, including changing dynamics, exploit determinism, and adding irrelevant confounders to high-dimensional observation.
An agent trained in these setups can overfit any of these factors, resulting in poor generalization performance. 
Many approaches have been proposed to address this challenges including various data augmentation approaches such as random cropping, adding jitter in image-based observation 
\citep{cobbe2019quantifying,laskin2020reinforcement,raileanu2020automatic,kostrikov2020image,laskin2020curl},random noise injection \citep{igl2019generalization}, network randomization \citep{osband2018randomized,burda2018exploration,Lee2020Network}, and regularization \citep{cobbe2019quantifying,kostrikov2020image,igl2019generalization,wang2020improving} have shown to improve generalization.
The common theme of these approaches is to increase diversity in the training data so as the learned policy would better generalize. 
However, this perturbation is primarily done in isolation of the RL reward function, which might change an essential aspect of the observation, resulting in sub-optimal policy learning. Moreover, the random perturbation in various manipulations of observations such as cropping, blocking, or combining two random images from different environment levels might result in unrealistic observations that the agent will less likely observe during testing. Thus these techniques might work poorly in the setup where agents depend on realistic observation for policy learning. It is also desirable to train the agent with realistic observations, which helps it understand the environments' semantics. Otherwise, the agent might learn unexpected and unsafe behaviors while entirely focusing on maximizing rewards even by exploiting flaws in environments such as imperfect reward design. 

This paper focuses on a particular form of generalization where the agent gets to observe a high-dimensional projection of the true state. In contrast, the latent state dynamics remain the same across all the MDP levels. 
This assumption has been shown to affect generalization \citep{Song2020Observational} in well studied benchmarks \citep{nichol2018gotta}. It has been observed that the agent overfits the scoreboard or timer and sometimes achieves the best training performance even without looking at the other part of the observation. Consequently, this policy completely fails to generalize in test environments. The high-capacity model even can memorize the whole training environment, thus severely affecting the generalization \citep{zhang2018study}.

This paper proposes a style transfer-based observation translation method that considers the observation's content and reward signal. A generator performs the style transfer while the policy tries to be robust toward such style changes. 
In particular, we propose a unified learning approach where the style transfer is integrated into the reinforcement learning training. Thus the generator produces more effective translation targeted toward RL objective, making the RL policy more robust. 

In our setup, the trajectory data from the agent's replay buffer is first clustered into different categories, and then observation is translated from one cluster's style to another cluster's style. Here the style is determined by the attribute commonality of observation features in a cluster. 
The agent should be robust toward changes of such features. Moreover, the translated trajectories correspond to those that possibly appear in testing environments, assisting the agent in adapting to unseen scenarios. 

Figure \ref{fig:framework} shows an overview of our method.
Our approach consists of a max-min game theoretic objective where a generator network modifies the observation by style transferring it to maximize the agent's probability of taking a different action. In contrast, the policy network (agent) updates its parameters to minimize the effect of perturbation while maximizing the expected future reward. Based on this objective, we propose a practical deep reinforcement learning algorithm, Adversarial Robust Policy Optimization (ARPO), that aims to find a robust policy by mitigating the observational overfitting and eventually generalizes to test environments.

We empirically show the effectiveness of our ARPO agent in generalization and sample efficiency on challenging Procgen \citep{cobbe2020leveraging} and Distracting Control Suite \citep{stone2021distracting} benchmark with image-based observation. The Procgen leverages procedural generation to create diverse environments to test the RL agents' generalization capacity.
The distracting control adds various visual distractions on top of the Deepmind Control suite \citep{tassa2020dmcontrol} to facilitate the evaluation of generalization performance.

Empirically, we observed that our agent ARPO performs better in generalization and sample efficiency than the baseline PPO \citep{schulman2017proximal} and a data augmentation-based RAD \citep{laskin2020reinforcement} algorithm on the many Procgen environments. Furthermore, APRO performs better compared to PPO and SAC \citep{haarnoja2018soft} on the two Distracting Control Suite environments in various setups. 

In summary, our contributions are listed as follows:
\begin{itemize}
    \item We propose Adversarial Robust Policy Optimization (ARPO), a deep reinforcement learning algorithm to find a robust policy that generalizes to test environments in a zero-shot generalization framework.
    
    \item We evaluate our approach on challenging Procgen \citep{cobbe2020leveraging} and Distracting Control Suite \citep{stone2021distracting} benchmark in generalization ans sample efficiency settings. 

    \item Empirically, we observed that our agent ARPO performs better generalization and sample efficiency performance compared to standard PPO \citep{schulman2017proximal}, SAC \citep{haarnoja2018soft}, and a data augmentation-based RAD \citep{laskin2020reinforcement} approach in various settings.
\end{itemize}

%% file: background.tex
\section{Preliminaries and Problem Settings} \label{sec:background}
\noindent\textbf{Markov Decision Process (MDP)} 
An MDP is denoted by $\mathcal{M} =(\mathcal{S}, \mathcal{A}, \mathcal{P}, r)$ where at every timestep $t$, from an state $s_t \in \mathcal{S}$ , the agent takes an action $a_t$ from a set of actions $\mathcal{A}$. The agent then receives a reward $r_t$ from the environment and move to a new state $s_{t+1} \in \mathcal{S}$ based on the transition probability $P(s_{t+1}|s_t, a_t)$. 

\noindent\textbf{Reinforcement Learning}.
Reinforcement learning aims to learn a policy $\pi \in \Pi$ that maps state to actions. The policy's objective is to maximize cumulative reward in an MDP, where $\Pi$ is the set of all possible policies. Thus, the policy which achieves the highest possible cumulative reward is the optimal policy $\pi^* \in \Pi$.

\noindent\textbf{Generalization in Reinforcement Learning}.
In the zero-shot generalization framework  \citep{Song2020Observational}, we assume the existence of a distribution of levels $\mathcal{D}$ over an MDP and a fixed optimal policy $\pi^*$ which can achieve maximal return over levels generated from the distribution $\mathcal{D}$. The levels may differ in observational variety, such as different background colors. In this setup, the agent has access to a fixed set of MDP levels during training the policy. The trained agent is then tested on the unseen levels to measure the generalization performance of the agent.
This scenario is often called Contextual MDP (i.e., CMDP). A detailed analysis can be found in the survey paper \cite{kirk2021survey}.

\noindent\textbf{Style Transfer}.
The task of image-to-image translation is to change a particular aspect of a given image to another, such as red background to green background. Generative adversarial network (GAN)-based method has achieved tremendous success in this task \citep{kim2017learning,isola2017image,zhu2017unpaired,choi2018stargan}.
Given training data from two domains, these models learn to style-translate images from one domain to another.  The domain is defined as a set of images sharing the same attribute value, such as similar background color and texture. The shared attributes in a domain are considered as the ``style'' of that domain. Many translation methods require paired images from two domains, which is not feasible in the reinforcement learning setup. The agent collects the data during policy learning, and annotating them in pairs is not feasible. Thus, we leverage unpaired image to image translation, which does not require a one-to-one mapping of annotated images from two domains. 

In particular, we build upon the work of StarGAN \citep{choi2018stargan} which is capable of learning mappings among multiple domains efficiently. The model takes in training data of multiple domains and learns the mappings between each pair of available domains using only a single generator. 
This method automatically captures special characteristics of one image domain and figures out how these characteristics could be translated into the other image collection, making it appealing in the reinforcement learning setup. However, the RL agent generates experience trajectory data which is not separated into domains. Thus, we further apply a clustering approach which first clusters the trajectory observation into domains, and then we train the generator, which style-translates among those domains.

%% file: method.tex
\section{Adversarial Robust Policy Optimization (ARPO)}
Our approach consists of a max-min game theoretic objective where a generator network modifies the observation by changing the style of it to maximize the agent's probability of taking a different action. In contrast, the policy network (agent) updates its parameters to minimize the effect of perturbation while maximizing the expected future reward. Based on this objective, we propose a practical deep reinforcement learning algorithm, Adversarial Robust Policy Optimization (ARPO), to find a robust policy that generalizes to test environments.
We now discuss the details of the policy network and generator.

\subsection{Policy Network}
The objective of the policy network is two-fold: maximize the cumulative RL return during training and be robust to the perturbation on the observation imposed by the generator. Ideally, the action probability of the translated image should not change from the original image, as the translated observation is assumed to represent the same semantic as the original observation.

The reinforcement learning loss  $\mathcal{L}_\pi$ can be optimized with various reinforcement learning mechanisms, such as policy gradient (optimization). However, we do not make any assumption the type of RL algorithms to use for optimizing $\mathcal{L}_\pi$; thus, any algorithms can be incorporated with our method.
Here we discuss the policy optimization formulation, which is based on the proximal policy optimization (PPO) \citep{schulman2017proximal} objective:
\begin{equation}\label{eq:policy_rl}
   \mathcal{L}_\pi = - \mathbb{E}_t[\frac{\pi_\theta(a_t \vert s_t)}{\pi_{\theta_{old}}(a_t \vert s_t)} A_t]
\end{equation}
\begin{equation}\label{eq:policy_rl_adv}
    A_t = -V(s_t) + r_t + \gamma r_{r+1} + ... + \gamma^{T-t+1} r_{T-1} + \gamma^{T-t} V(s_T).
\end{equation}
where, $\pi_{\theta}$ is the current policy and $\pi_{\theta_{old}}$ is the old policy; $A_t$ is estimated from the sampled trajectory by policy $\pi_{\theta_{old}}$ leveraging a value function estimator denoted as $V$. Here, both the policy $\pi$ and value network $V$ are represented as neural networks.

An additional objective of the policy is to participate in the adversarial game with the style transfer generator. Thus the final policy loss $\mathcal{L}_\theta$ is defined as follows:
\begin{equation}\label{eq:policy}
    \mathcal{L_\theta} =  \mathcal{L}_\pi + \beta_1 KL[\pi_\theta(. \vert x_t), \pi_\theta(. \vert x_t')] 
\end{equation}

The policy parameter $\theta$ is updated to minimize equation \ref{eq:policy}. For the observation $x_t$ at timestep $t$, the KL-component measure the distortion in the action distribution of the policy $\pi_\theta$ due to the perturbation ($x_t'$).
The hyperparameter $\beta_1$ determines the participation in an adversarial objective with the generator. By minimizing the KL-part, the policy becomes robust to the change in perturbation proposed by the generator.

\subsection{Generator Network}
It takes the observation $x_t$ as input and outputs the style-translated observation $x_t'$. The objective of this network is to change the agent's (policy network's) probability of taking a different action for the corresponding input observation. However, the content or semantic of the observations needs to be intact, and the only style of the images will differ. 

\begin{figure}[h] 
    \centering
    \includegraphics[width=0.7\linewidth]{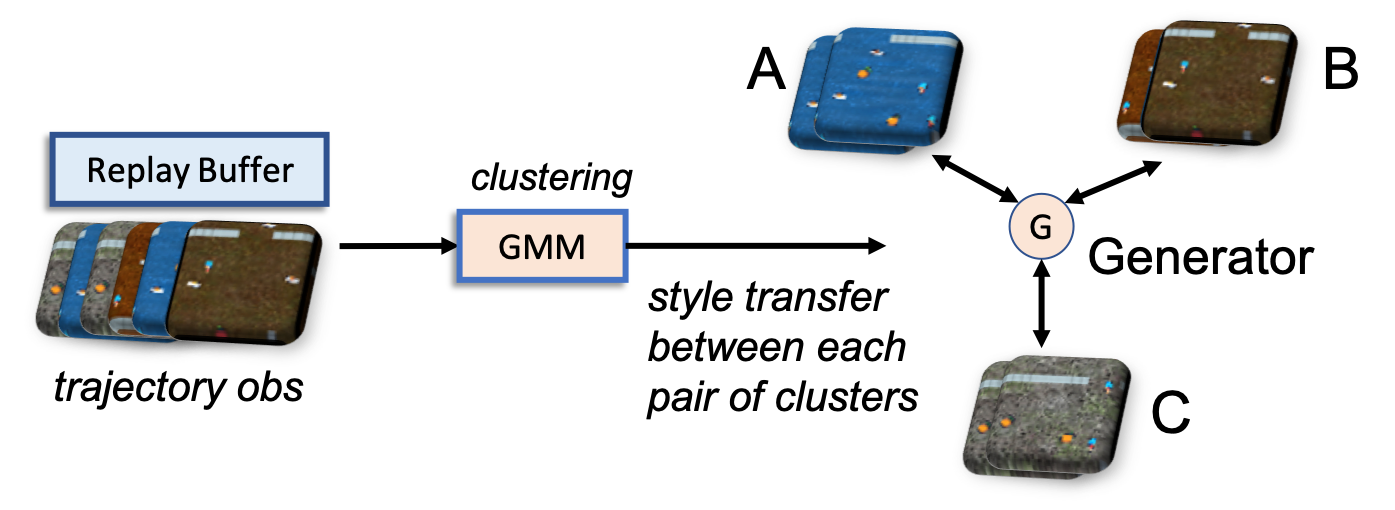}
    \caption{\small  Overview of style transfer generator module. The experience trajectory for the task environment are separated into different classes based on visual features using Gaussian Mixture Model (GMM) clustering algorithm. The task of the generator is then to translate the image from one class image to another classes images. In this case, the ``style" is defined as the commonality among images in a single class. Given a new observation, it first infers into its (source) cluster using GMM, and then the generator translated it to (target) another cluster style. The target cluster is taken randomly from the rest of the cluster.
}
    \label{fig:generator_framework}
\end{figure}

The first step of this process is to curate a dataset with different style features such as background color and texture of objects present in the observation. We first use experience trajectory observation data from the environment and separate them into different classes based on visual features (Figure \ref{fig:generator_framework}). The task of the generator is then to translate the images from one class images to another class images. In this case, the ``style" is defined as the commonality among images in a single class.
We now discuss details about how we separate the trajectory observation to get different clusters. 

\noindent\textbf{Clustering Trajectory Observation}.
The trajectory data is first clustered into $n$ clusters. Then, we use the Gaussian Mixture Model (GMM) for clustering and ResNet \citep{he2016deep} for feature extraction and dimension reduction. 
Firstly, a pre-trained (on ImageNet) ResNet model reduces an RGB image into a 1000 dimensional vector. This step extracts useful information from the raw image and reduces the dimension, allowing faster cluster training. Note that this process focuses entirely on the visual aspect of the observation. 
After passing through the ResNet, the observation dataset is clustered using GMM.

\textbf{Generator Training}.
Generator $G$ tries to fool discriminator $D$ in an adversarial setting by generating a realistic image represented by the true image distribution. We extend the generator's ability, and an additional objective of the generator is to fool the RL agent (policy network) by transforming the style of the original observation. For example, changing the background color to a different one does not change the semantic of the observation, but the image style would be different. Ideally, the action mapping of the policy needs to be the same on both the original and translated observation.
However, a policy that focuses on the irrelevant background color will suffer from this translation, and thus its probability distribution deviates because of the style translation. This is defined as the KL-divergence $KL(\pi_\theta(.|x_t), \pi_\theta(.|x_t')) $, where $x_t$ is the given observation and $x_t'$ is the translated observation by the generator. 

The generator tries to generate a style that eventually increases the KL quantity. On the other hand, the policy should learn to be robust toward such changes. The policy training described above addresses this and tries to minimize the KL-divergence due to the style transfer, which eventually allows robust policy learning and ultimately prevents the agent from overfitting irrelevant features from the high-dimensional observation.

We build on the generator on previous works \citep{choi2018stargan,zhu2017unpaired} which is a unified framework for a multi-domain image to image translation.
The discriminator loss is
\begin{equation} \label{eq:loss_disciminator}
    \mathcal{L}_D = -\mathcal{L}_{adv} + \lambda_{cls} \mathcal{L}^r_{cls},
\end{equation}
which consist of the adversarial loss $\mathcal{L}_{adv}$ and domain classification loss $\mathcal{L}^r_{cls}$. The discriminator detects a fake image generated by the generator G from the real image in the given class data.

On the other hand, the generator loss is
\begin{equation} \label{eq:loss_generator}
    \mathcal{L}_G = \mathcal{L}_{adv} + \lambda_{cls} \mathcal{L}^f_{cls} +\lambda_{rec} \mathcal{L}_{rec} - \beta_2 KL(\pi_\theta(.|x_t), \pi_\theta(.|x_t')) 
\end{equation}
which consists of image translation components and policy component which is the KL-part. From the translation side, the $\mathcal{L}_{adv}$ is adversarial loss with discriminator $\mathcal{L}^f_{cls}$ is the loss of detecting fake image. 
Finally, a cycle consistency loss 
 $\mathcal{L}_{rec}$ \citep{choi2018stargan,zhu2017unpaired} is used which makes sure the translated input can be translated back to the original input, thus only changing the domain related part and not the semantic. 
Some more details of these losses can be found in the supplementary materials.

Our addition to the generator objective is the KL-part in equation \ref{eq:loss_generator} which tries to challenge the RL agent and fool its policy by providing a more challenging translation. 
However, the eventual goal of this adversarial game is to help the agent to be robust to any irrelevant features in the observations.
The hyperparameter $\beta_2$ control the participation in the adversarial objective of the generator network. 

Now we discuss a practical RL algorithm based on the above objectives.

\subsection{ARPO Algorithm}
The training of the policy network and the generator network happen in an alternating way. First, the policy is trained on a batch of replay buffer trajectory observation, and then the generator parameters are updated based on the objectives discussed above. The more overall training progresses, the better the generator at translating observation, and the policy also becomes better and becomes robust to irrelevant style changes. For stable training, the input images data for the generator is kept fixed initially, and we apply the GMM cluster only at the beginning to generate different class images (distribution). The pseudocode of the algorithm is given in Algorithm \ref{algo-arpo}.

\begin{algorithm}
\caption{Adversarial Robust Policy Optimization (ARPO)}
\label{algo-arpo}
\begin{algorithmic}[1]
    \STATE Initialize parameter vectors for policy network and generator network
    \FOR {each iteration}
    
        \FOR{each environment step} 
            \STATE $a_t \sim \pi_\theta(a_t|x_t)$ 
            \STATE $x_{t+1} \sim P(x_{t+1}|x_t, a_t)$ 
            \STATE $r_t \sim R(x_t, a_t)$ 
            \STATE $\mathcal{D} \xleftarrow{} \mathcal{D} \cup \{(x_t, a_t, r_t, x_{t+1})\} $ 
        \ENDFOR
        
        \FOR{each observation $x_t$ in $\mathcal{D}$}
        \STATE $x_t' \xleftarrow{} Generator(x_t)$ \textit{// generate translated observation}
        \STATE Compute $\mathcal{L}_\pi$ from data $\mathcal{D}$ using Equation \ref{eq:policy_rl}, and \ref{eq:policy_rl_adv} \textit{// update policy for RL objective}
        \STATE Compute $\mathcal{L}_\theta$ using Equation \ref{eq:policy} \textit{// update policy with adversarial KL part}
        \STATE Update generator by computing Equation \ref{eq:loss_disciminator}, and \ref{eq:loss_generator} 
        \ENDFOR
    \ENDFOR
\end{algorithmic}
\end{algorithm}

\textbf{Discussion on convergence}.
The adversarial min-max KL component $KL(\pi_\theta(.|x_t), \pi_\theta(.|x_t'))$ become zero when the RL policy is optimal ($\pi^*$) and the generators only perturbs (translates) the irrelevant part of all observations. In that case, the optimal policy only focuses on the relevant part of the observations, that is, the true state; thus, any changes in irrelevant part due to style transfer will be ignored. At that point the KL component become zero as the $\pi^*_\theta(.|x_t) = \pi^*_\theta(.|x_t')$. Note that in practice, the algorithm is usually trained for limited timesteps, and thus the adversarial training might not converge to optimal. However, empirically we observe that our algorithm ARPO achieves improved performance in many challenging Procgen and Distracting control suite tasks.

%% file: experiment.tex
\section{Experiments} \label{sec:exp}
\subsection{Setup} 
Our experiments cover both discrete (Procgen) and continuous (Distracting control) action spaces with image-based observation. 

\begin{figure}[!ht]
\centering
\includegraphics[width=0.60\linewidth]{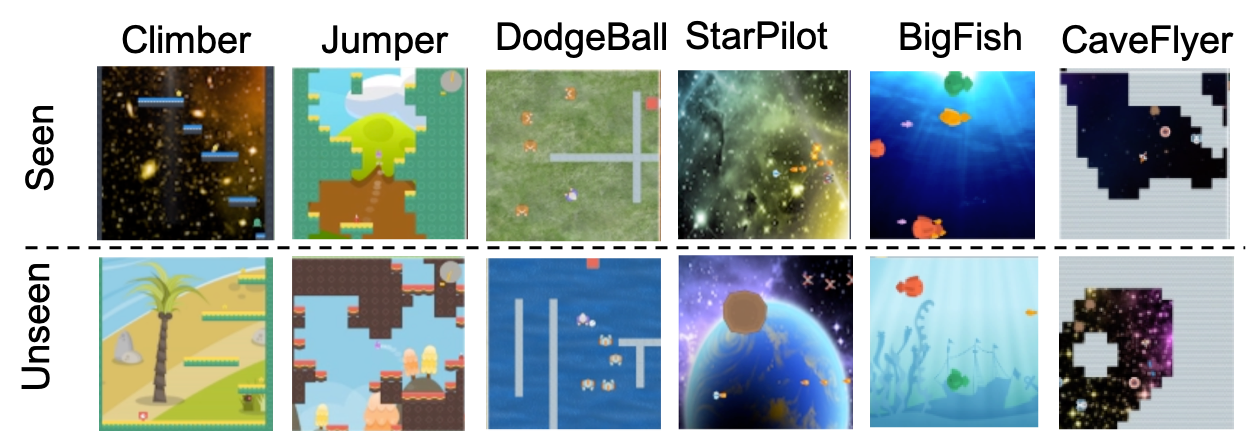} 
\caption{\small  
Some snippets of different Procgen environments \citep{cobbe2020leveraging}. The training (seen) levels vary drastically from the testing (unseen) environment. The agent must master the skill without overfitting irrelevant non-generalizable aspects of the environment to perform better in unseen levels.
}
\centering
\label{fig:procgen_env}
\end{figure}

\noindent\textbf{Procgen Environment}.
We evaluate our proposed agent ARPO on the challenging RL generalization benchmark Procgen \citep{cobbe2020leveraging}.
We use the standard evaluation protocol from \cite{cobbe2020leveraging} where the agent is allowed to train on only 200 levels of each environment, with the difficulty level set as \textit{easy}. Then the agents are tested over more than 100K levels, full distribution. Thus, to better generalize to the test environment, the agent must master the skill and avoid overfitting to spurious features in the training environment. This setup is in-distribution generalization as a fixed number of train levels are randomly taken from an i.i.d level distribution.
The observations are raw pixel images, and the environment varies drastically between levels; some snippets are given in Figure \ref{fig:procgen_env}. 

\noindent\textbf{Distracting Control Suite Environment}.
\begin{figure*}
\centering
\includegraphics[width=0.7\linewidth]{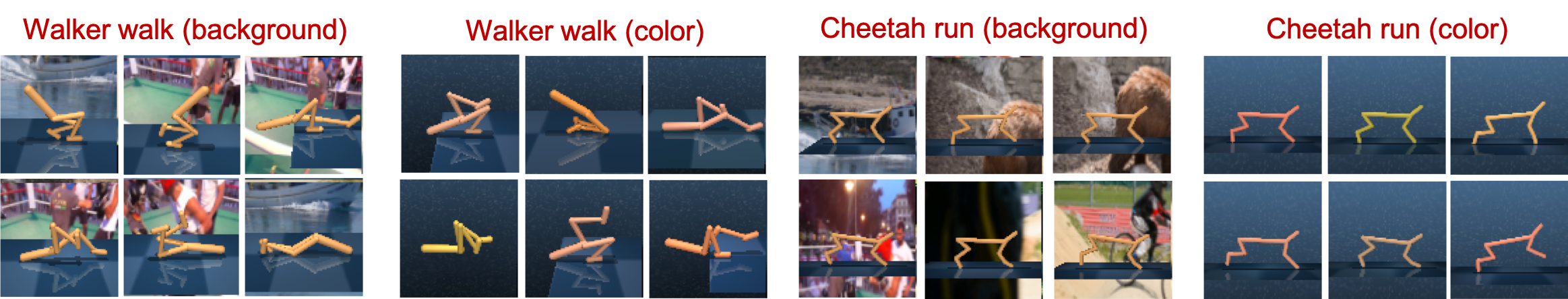} 
\caption{\small  
Some snippets from Distracting control suite \citep{stone2021distracting}. 
}
\centering
\label{fig:ddmc_env}
\end{figure*}
The distracting control adds various visual distractions on top of the Deepmind Control suite \citep{tassa2020dmcontrol} to facilitate the evaluation of generalization performance. We experimented on two distracting settings, \textit{background (dynamic)} and \textit{color} for \textit{easy} difficulty mode. In the distracting background, videos played in the background, and for the color distraction, the body of the robot color changes in different agent interactions. The train and test environments differ by different distractions, such as separate videos for train and test. As the agent has to use image-based observation, the distractions make the task challenging. This setup is an out-of-distribution generalization as the test time videos differ from train videos and are unseen during training.
Some snippets of the environments can be found in Figure \ref{fig:ddmc_env}.

\noindent\textbf{Baselines}.
We compare our algorithm with Proximal Policy Optimization (\textbf{PPO}), an on-policy algorithm motivated by how we can quickly optimize our policy while staying close to the current policy. PPO performs effectively in many RL benchmarks, which include the Procgen environments \citep{cobbe2020leveraging}.
Furthermore, we evaluated our method with a data augmentation technique, \textbf{RAD} \citep{laskin2020reinforcement} on Procgen environments. RAD is a plug-and-play module that can be used with existing RL algorithms and shows effective empirical performance in complex RL benchmarks, including some Procgen environments. In particular, we evaluated \textit{Cutout Color} augmentation technique which has shown better results in many Procgen environments compared to other augmentation techniques evaluated in \cite{laskin2020reinforcement}. This augmentation is applied on top of standard PPO. Note that our ARPO algorithm can be easily plugged in with these data augmentation-based methods. However, in this paper, we evaluated how our ARPO agent performs even when no augmentation is used.
Furthermore, in the Distracting control suite, in addition to PPO, we compare ARPO with Soft Actor-Critic (\textbf{SAC}) \citep{haarnoja2018soft} which is an off-policy algorithm that optimizes a stochastic policy.  

\noindent\textbf{Implementation}.
We implemented our ARPO on the Ray RLlib framework \citep{liang2018rllib} and used a CNN-based model for the policy network for these experiments.  
For all the experiments with ARPO, we set the $\beta_1=\beta_2=20$ (in equation \ref{eq:policy}, and \ref{eq:loss_generator}). The generator's hyperparameters are set to $\lambda_{cls} = 1$, and $\lambda_{rec} = 10$. Number of cluster is set to 3.
Details of model configurations and parameters are given in the Appendix.

\subsection{Procgen Results}
\textbf{Sample Efficiency}.
We evaluated and compared our agent on the sample efficiency during training. This result shows how quickly our agent achieves a better score during training. Table \ref{tab:comparison_results_procgen_train} shows the results after training for 20 million timesteps. Learning curve can be found in Appendix.

\begin{table}
\caption{[\textbf{Train Reward}] Sample efficiency results on Procgen environments. The results are the after training agents for 20M timesteps. The mean value and standard deviation are calculated across 3 random seed runs. ARPO achieves the best generalization results in many environments compared to PPO and RAD. Best agent is in \textbf{Bold}. 
} 
\label{tab:comparison_results_procgen_train} 
\begin{center}

\begin{tabular}{|c|c|c|c|c|}
 \hline
 \textbf{Env} &\textbf{ ARPO (ours)} & \textbf{PPO}  & \textbf{RAD}\\
\hline
jumper & \textbf{8.46} $\pm$ \small{0.22} & 8.41 $\pm$ \small{0.07} & 7.94 $\pm$ \small{0.47} \\
\hline
bigfish & \textbf{10.69} $\pm$ \small{2.26} & 8.19 $\pm$ \small{1.96} & 7.48 $\pm$ \small{0.61} \\
\hline
climber & \textbf{7.56} $\pm$ \small{0.58} & 6.85 $\pm$ \small{0.92} & 6.23 $\pm$ \small{0.63} \\
\hline
caveflyer & \textbf{6.78} $\pm$ \small{0.48} & 6.14 $\pm$ \small{1.05} & 5.12 $\pm$ \small{0.49} \\
\hline
miner & \textbf{10.62} $\pm$ \small{0.83} & 7.87 $\pm$ \small{1.13} & 6.59 $\pm$ \small{1.16} \\
\hline
fruitbot & \textbf{27.29} $\pm$ \small{0.68} & 27.26 $\pm$ \small{0.96} & 26.88 $\pm$ \small{0.35} \\
\hline
leaper & \textbf{4.47} $\pm$ \small{0.35} & 3.98 $\pm$ \small{1.06} & 2.72 $\pm$ \small{0.51} \\
\hline
ninja & 8.07 $\pm$ \small{0.35} & \textbf{8.08} $\pm$ \small{0.78} & 6.11 $\pm$ \small{0.82} \\
\hline
dodgeball & \textbf{5.34} $\pm$ \small{0.4} & 5.13 $\pm$ \small{0.51} & 3.8 $\pm$ \small{0.31} \\
\hline
maze & \textbf{9.09} $\pm$ \small{0.18} & 7.77 $\pm$ \small{0.59} & 6.26 $\pm$ \small{0.42} \\
\hline
plunder & 8.66 $\pm$ \small{1.05} & \textbf{9.55} $\pm$ \small{1.03} & 8.14 $\pm$ \small{2.0} \\
\hline
starpilot & 24.61 $\pm$ \small{3.16} & 26.17 $\pm$ \small{1.22} & \textbf{27.36} $\pm$ \small{3.13} \\
\hline
heist & \textbf{5.1} $\pm$ \small{0.26} & 4.7 $\pm$ \small{0.44} & 3.93 $\pm$ \small{0.5} \\
\hline
coinrun & 8.54 $\pm$ \small{0.68} & \textbf{9.57 }$\pm$ \small{0.09} & 9.15 $\pm$ \small{0.24} \\
\hline
bossfight & 4.11 $\pm$ \small{1.04} & 6.87 $\pm$ \small{0.56} & \textbf{7.97} $\pm$ \small{0.82} \\
\hline
chaser & 2.51 $\pm$ \small{0.17} & \textbf{2.82} $\pm$ \small{0.55} & 2.5 $\pm$ \small{0.27} \\
\hline
\end{tabular}
\end{center}
\end{table}

In many environments, we see our agent ARPO perform better than the baselines PPO and RAD. These results show that despite optimizing for the generalization, our adversarial training helps the ARPO agent learn a robust policy which is also better during training.

\begin{figure*}[!ht]
\centering
\includegraphics[width=0.99\linewidth]{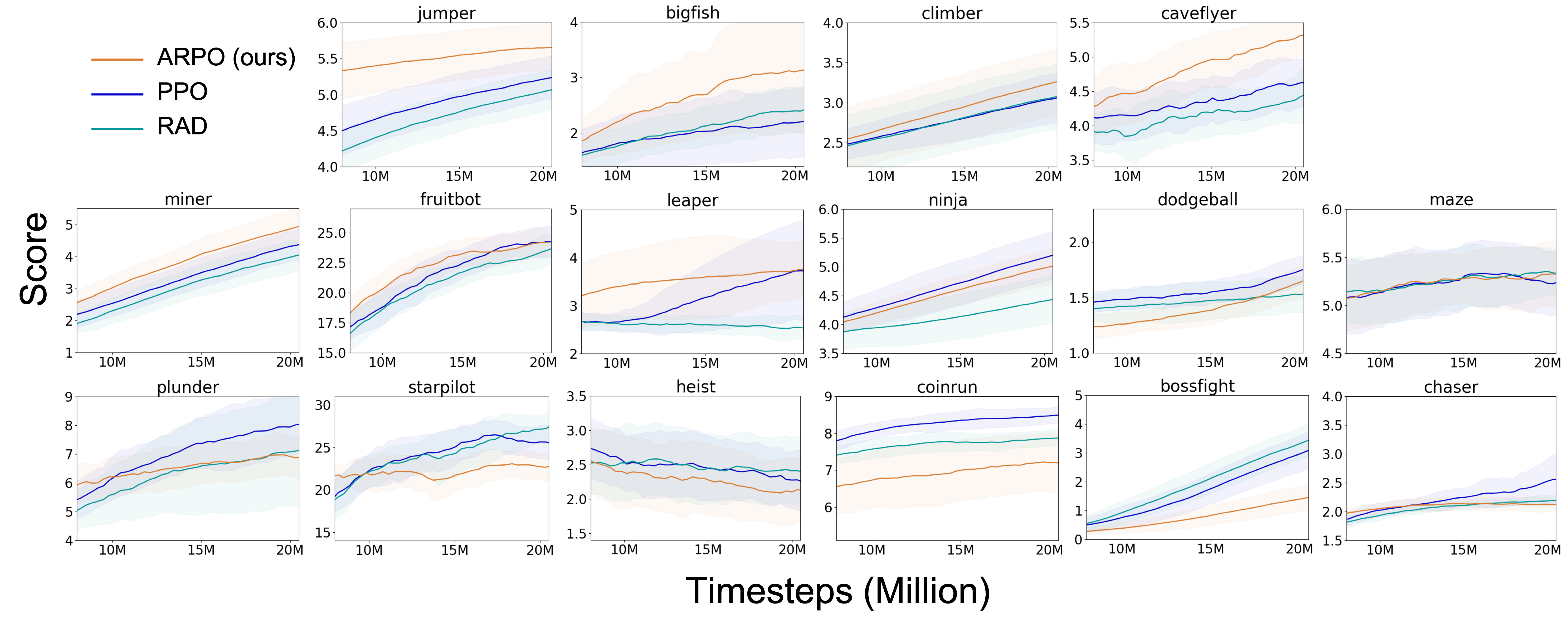} 
\caption{\small  
Generalization result learning curve. Test performance comparison on Procgen environments with image observation. ARPO performs better in many environments where it outperforms the baselines. The results are averaged over 3 seeds.
}
\centering
\label{fig:arpo_procgen_test}
\end{figure*}

\textbf{Generalization}.
The results in Figure \ref{fig:arpo_procgen_test}  show the test results of ARPO and baselines PPO and RAD on 16 Procgen environments with image observation. Overall, ARPO performs better in many environments. We observe that, almost in half of the environments, our agent ARPO outperforms both baselines. It also performs comparably in a few other environments.
Note that in some environments such as Ninja, Plunder, Chaser, Heist, all the agents fail to obtain a reasonable training accuracy \citep{cobbe2020leveraging} and thus perform poorly during testing (generalization). 
This result shows that our proposed agent ARPO achieves a strong performance and reduces the generalization gap compared to the tested baselines on generalization.

We further show how each agent achieves generalization during the entire timesteps. Consequently, we compute the reward score achieved  by each agent after training for 20 million timesteps. Finally we report mean and standard deviation of these scores across 3 seed runs. The generalization results are computed by evaluating the trained agents on test levels (full distribution). The performance is computed by evaluating each agent for 10 random episode trials after a training interval of 10.
Table \ref{tab:comparison_results_procgen_test} shows the results comparison for all the agents on  Procgen environments. Similar to the previous discussion, we see that in many of the environments, our ARPO agents perform better than the baselines. ARPO performs better than both the baselines in many environments. 
\begin{table}
\caption{[\textbf{Test Reward}] Generalization results on Procgen environments. The results are the after training agents for 20M timesteps. The mean value and standard deviation are calculated across 3 random seed runs. ARPO achieves the best generalization results in many environments compared to PPO and RAD. Best agent is in \textbf{Bold}. 
} 
\label{tab:comparison_results_procgen_test} 
\begin{center}
\begin{tabular}{|c|c|c|c|c|}
 \hline
 \textbf{Env} &\textbf{ ARPO (ours)} & \textbf{PPO}  & \textbf{RAD}\\
\hline
jumper & 5.65 $\pm$ \small{0.34} & \textbf{5.96} $\pm$ \small{0.05} & 5.95 $\pm$ \small{0.28} \\
\hline
bigfish & \textbf{2.92} $\pm$ \small{0.73} & 2.24 $\pm$ \small{0.57} & 2.48 $\pm$ \small{0.55} \\
\hline
climber & \textbf{4.73} $\pm$ \small{0.47} & 4.37 $\pm$ \small{0.44} & 3.95 $\pm$ \small{0.57} \\
\hline
caveflyer & \textbf{5.46} $\pm$ \small{0.21} & 4.77 $\pm$ \small{0.13} & 4.54 $\pm$ \small{0.52} \\
\hline
miner & \textbf{6.45} $\pm$ \small{0.33} & 5.95 $\pm$ \small{0.38} & 5.4 $\pm$ \small{0.33} \\
\hline
fruitbot & 24.51 $\pm$ \small{0.29} & 24.4 $\pm$ \small{1.57} & \textbf{25.02} $\pm$ \small{1.98} \\
\hline
leaper & 3.96 $\pm$ \small{0.47} & \textbf{4.14} $\pm$ \small{1.13} & 2.71 $\pm$ \small{0.25} \\
\hline
ninja & 5.89 $\pm$ \small{0.47} & \textbf{6.44} $\pm$ \small{0.59} & 5.06 $\pm$ \small{0.32} \\
\hline
dodgeball & \textbf{2.02} $\pm$ \small{0.18} & 2.0 $\pm$ \small{0.03} & 1.68 $\pm$ \small{0.25} \\
\hline
maze & 5.37 $\pm$ \small{0.41} & 4.79 $\pm$ \small{0.35} & \textbf{5.49} $\pm$ \small{0.03} \\
\hline
plunder & 6.38 $\pm$ \small{0.22} & \textbf{7.88} $\pm$ \small{0.83} & 7.22 $\pm$ \small{2.01} \\
\hline
starpilot & 21.86 $\pm$ \small{0.58} & 26.56 $\pm$ \small{4.0} & \textbf{27.28} $\pm$ \small{0.55} \\
\hline
heist & 1.9 $\pm$ \small{0.49} & \textbf{2.18} $\pm$ \small{0.57} & 2.16 $\pm$ \small{0.27} \\
\hline
coinrun & 7.01 $\pm$ \small{0.97} & \textbf{8.53} $\pm$ \small{0.15} & 7.93 $\pm$ \small{0.26} \\
\hline
bossfight & 2.88 $\pm$ \small{0.73} & 6.4 $\pm$ \small{1.3} & \textbf{7.17} $\pm$ \small{1.1} \\
\hline
chaser & 2.18 $\pm$ \small{0.02} & \textbf{2.85} $\pm$ \small{0.75} & 2.21 $\pm$ \small{0.06} \\
\hline
\end{tabular}
\end{center}
\end{table}

\subsection{Distracting Control Results}
\begin{figure*}[!ht]
\centering
\includegraphics[width=0.95\linewidth]{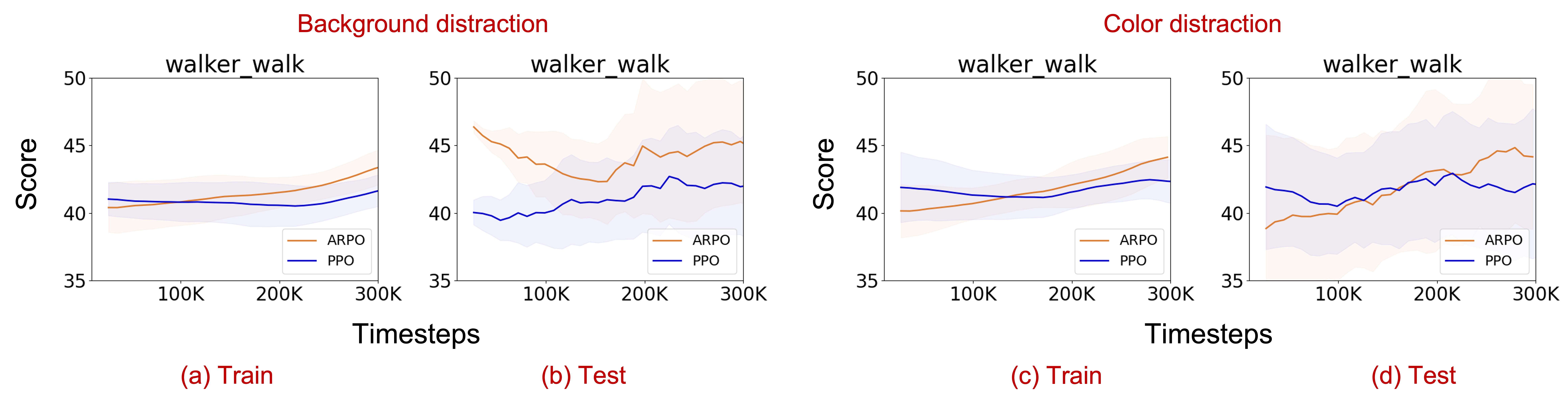} 
\caption{\small  
Sample efficiency (Train) and generalization (Test) results on Distracting control Walker walk environments. The results are averaged over 3 seeds.
}
\centering
\label{fig:ddmc_walker_walk}
\end{figure*}

Figure \ref{fig:ddmc_walker_walk} shows the results of ARPO and PPO on Walker walk environment from distracting control suite.
We observe that ARPO performs better compared to the baseline algorithm PPO in both sample efficiency (train) and generalization (test) for background and color distraction. Note that we observe large variance (standard deviation) in the results across the run, particularly in test time. These results correspond to the benchmark results where \cite{stone2021distracting} also find comparatively larger variance in reported results. In some cases (Figure \ref{fig:ddmc_walker_walk}b), the test performance surpasses the corresponding timestep's train reward. This scenario might happen because the videos for train and test environments have different difficulties. Thus in some cases, the agent finds better test performance than the training. However, our ARPO's performance remains better during testing compared to PPO. Further, ARPO (and PPO) performs better compared to the off-policy SAC in both sample efficiency and generalizations in our settings. Detailed results are in Appendix.

Furthermore, results analysis for Cheetah run environment is in the Appendix.

\subsection{Qualitative Analysis of Adversarial Style Transfer}
\begin{figure}[!ht]
\centering
\includegraphics[width=0.7\columnwidth]{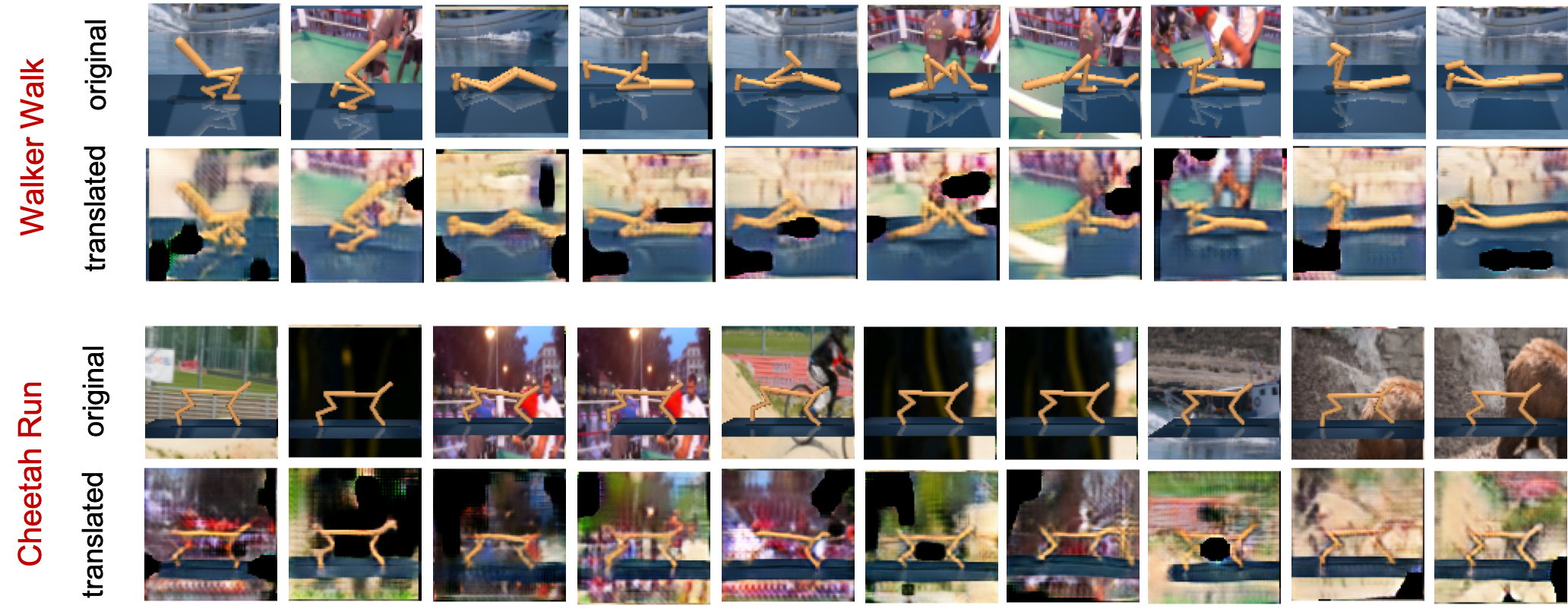} 
\caption{\small  
Sample translation of generator on \textbf{background} distraction for Walker-walk environment from Distracted Control Suite benchmark. We see that the translated observations retain the semantic that is the robot pose while changing non-essential parts. Interestingly, we observe that the adversarial generator blackout some non-essential parts of the translated images. This scenario might indicate that the min-max objective between policy network and generator tries to recover the actual state from the observation by removing parts irrelevant to the reward.
}
\centering
\label{fig:ddmc_translated_sample}
\end{figure}

\begin{figure}[!ht]
\centering
\includegraphics[width=0.70\columnwidth]{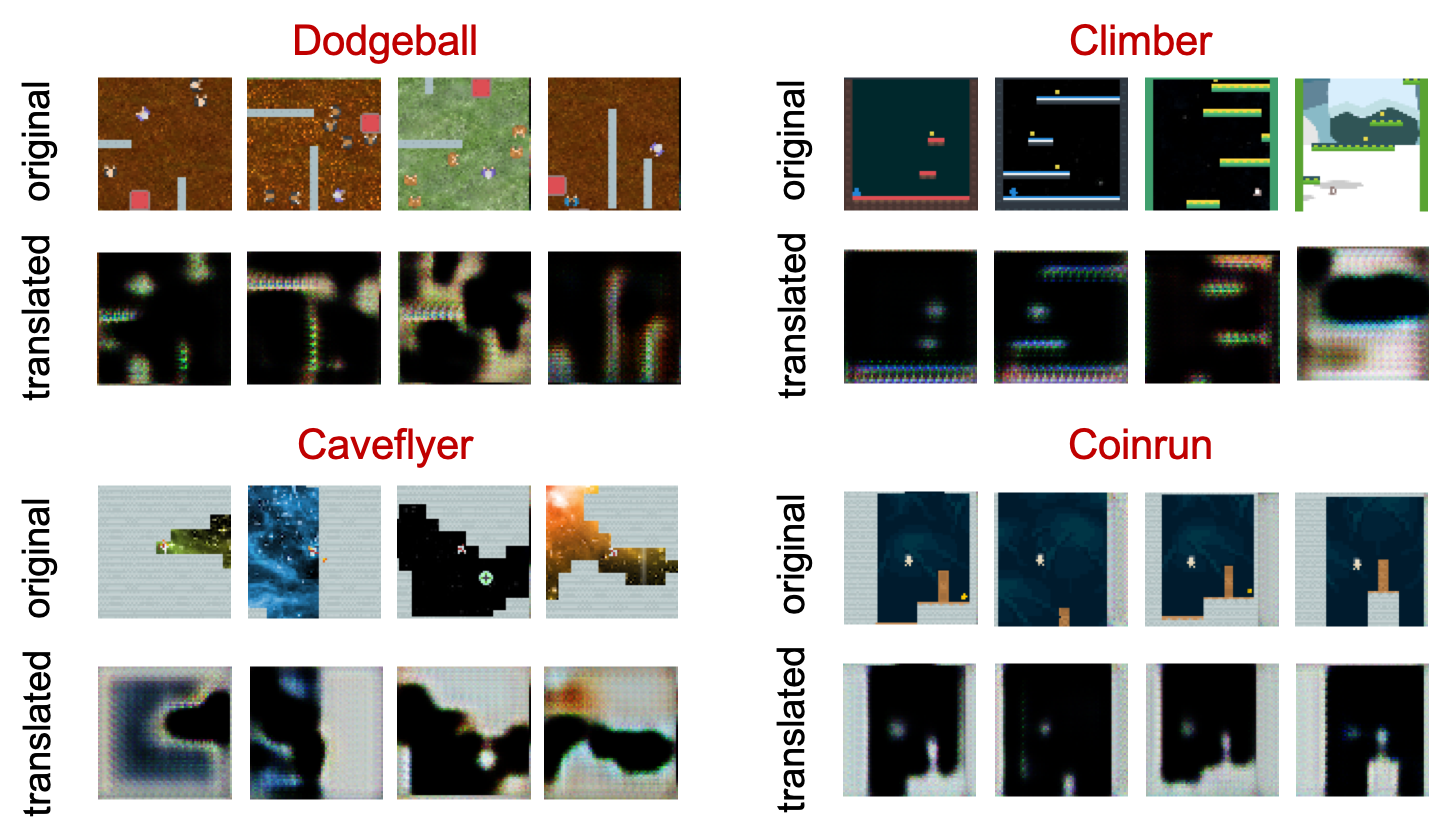} 
\caption{\small  
Sample translation of generator on four Procgen environments. We see that the generator retains most of the game semantic while changing the background color and the texture of various essential objects.
}
\centering
\label{fig:procgen_translated_sample}
\end{figure}

\begin{figure}[!ht]
\centering
\includegraphics[width=0.50\columnwidth]{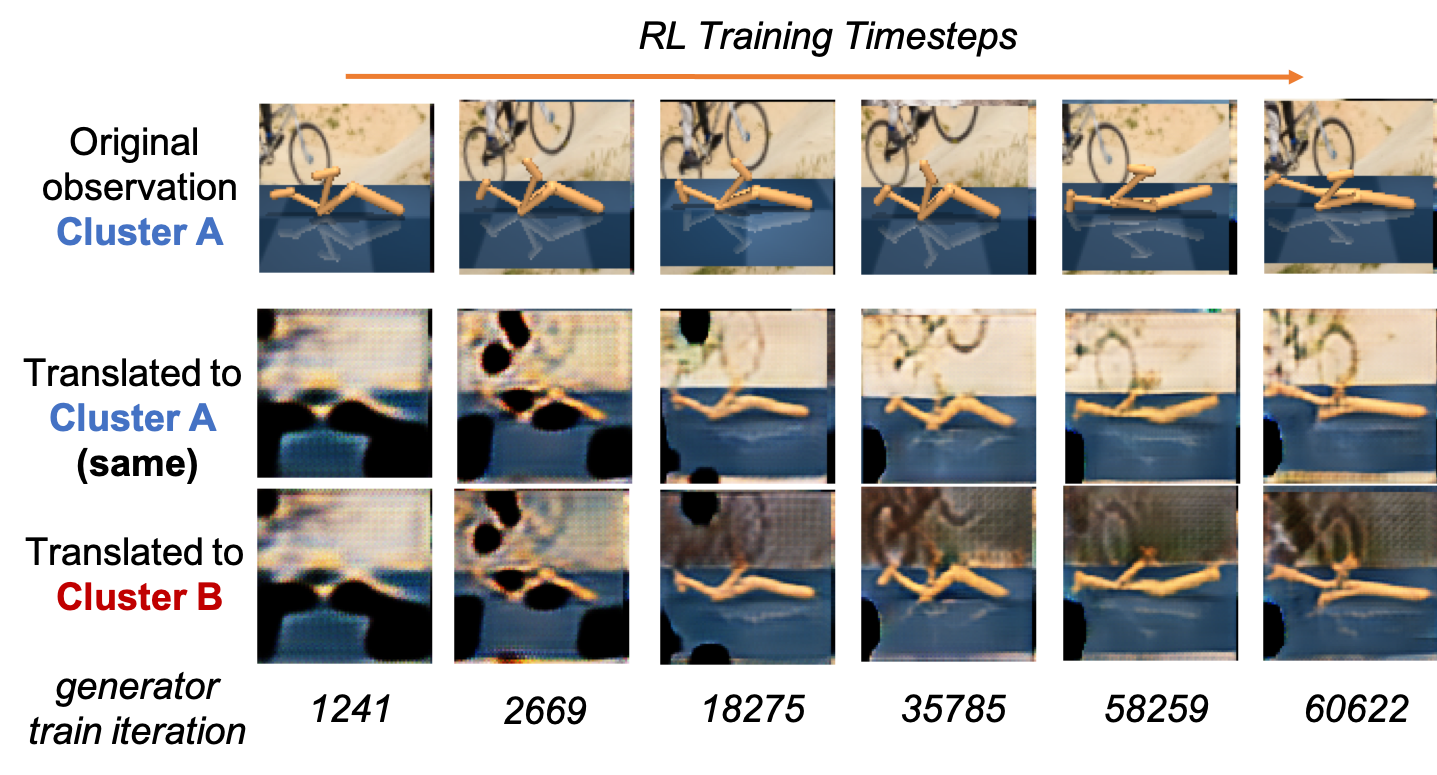} 
\caption{\small  
Sample translation of generator during various training phases on \textbf{background} distraction for Walker-walk environment from Distracted Control Suite benchmark. [\textbf{Second row}] We see that when translated back to the same cluster (in the second row), the generated images get some distortion (e.g., blackout some parts), while the essential parts remain mostly intact. This scenario might be happening due to the regularization of the cycle consistency loss by the KL component in equation \ref{eq:loss_generator}. [\textbf{Third row}] We see that the translation to a different cluster generates images that changes the irrelevant parts while mostly keeping the semantic intact. In both the second and third row, the translation gets better as time progresses, suggesting the adaptation of both policy and generator networks due to the min-max objective.
}
\centering
\label{fig:translation_sample_timeline}
\end{figure}

In this section we discuss how the generator performs in generating adversarially translated observation during agent training.
Figure \ref{fig:ddmc_translated_sample}, \ref{fig:procgen_translated_sample}, and \ref{fig:translation_sample_timeline} show sample translation from the generator. Please see the corresponding caption for detailed discussion. Some additional qualitative results are in the Appendix.

%% file: relatedwork.tex
\section{Related Work} 
\label{sec:related_work}
\textbf{Data augmentation in RL}.
In recent time, many approaches have been proposed to address generalization challenges including various data augmentation approaches \citep{cobbe2019quantifying,laskin2020reinforcement,kostrikov2020image,raileanu2020automatic,laskin2020curl,igl2019generalization,osband2018randomized,Lee2020Network,wang2020improving,zhang2021learning,rahman2023bootstrap} have shown to improve generalization.
The common theme of these approaches is to increase diversity in the training data; however, these perturbations are primarily performed in isolation of the RL reward function, which might change crucial aspects of the observation, resulting in sub-optimal policy learning. In contrast, we propose a style transfer that tries to generate semantically similar but different visual style observation to train a robust policy. Furthermore, our style perturbation takes into account the reward signal from the policy.

Note that the base RL algorithm (e.g., PPO) in our ARPO method still uses the original observation from the environment while participating in adversarial objectives (see Figure \ref{fig:framework}). Thus, data augmentation can be applied to the original observations before feeding them to the base RL algorithm to potentially improve the base RL agent, as observed in many data augmentation techniques discussed above. 

The experiment setup of SVEA \cite{hansen2021stabilizing} is different because they assume no visual distraction during training but distraction during test time. This scenario is in contrast to the benchmark we are evaluating. In this paper, we are also evaluating the existence of distraction during training.

\textbf{Style Transfer in RL}. 
Many approaches have been proposed which use style transfer to improve reinforcement learning algorithms.
\cite{gamrian2019transfer} use image-to-image translation between domains to learn the policy in zero-shot learning and imitation learning setup. They use different levels' data and train unpaired GANs between two levels (domains), which requires access to the annotated agent's trajectory data in both source and target domains.
Furthermore, \cite{smith2019avid} performs pixel-level image translation via CycleGAN to convert the human demonstration to a video of a robot. These translated data can then be used to convert the human demonstration into a video of a robot used to construct a reward function that helps learn complex tasks. However, their CycleGAN operates on two predefined domains, human demonstration and robot video. 

In contrast, we first automatically generate the domains using a clustering algorithm and then train the style-translation generator. In our case, we do not need the information of levels; instead, we automatically cluster the trajectory data based on visual features of the observations.  Additionally, our visual-based clustering may put observations from multiple levels into a cluster, potentially preventing GAN from overfitting \cite{gamrian2019transfer} to any particular environment levels.

\textbf{Adversarial Approach in RL}.
Adversarial methods have been used in the context of reinforcement learning to improve policy training.
\cite{li2021domain} proposes to combine data augmentation with auxiliary adversarial loss to improve generalization.  In particular, they use an adversarial discriminator to predict the label of the observation. However, they need to know the domain label, limiting the applicability as it might not be available, and the number of labels might be large, for example, in procedural generation. In contrast, we use an unsupervised clustering that automatically finds out similar visual features later used for generator training. 
\citep{pinto2017robust} use adversarial min-max training to learn robust policy against forces/disturbances in the system \citep{pinto2017robust}. In addition, adversarial training has been used between two robots to improve object grasping \citep{pinto2017supervision} and in the context of multi-agent self-play scenario \citep{heinrich2016deep}. 
\cite{zhang2021generalization} generate adversarial examples for observations in a trajectory by minimizing the cumulative reward objective of the reinforcement learning agent. In contrast to these methods, we use adversarial visual-based style-transfer of the observation to guide robust policy learning by enforcing the policy to produce similar output action distributions.

\textbf{Generalization in RL}.
Many methods have been proposed to reduce the generalization gap between training and unseen test environment in the context of reinforcement learning. State learning approaches \citep{higgins2017darla,agarwal2021pse,zhang2020learning}  and auto-encoder based latent state learning \citep{lange2012autonomous,lange2010deep,hafner2019learning} with reconstruction loss have been proposed. The sequential structure in RL \citep{agarwal2021pse} and behavioral similarity between states \citep{zhang2020learning} to learn invariant states have been leveraged to improve RL robustness and generalization.  In contrast, we learn the invariant state by providing the RL agent with an additional variant of the same observation while retaining the reward structure using an adversarial max-min objective.

Our primary focus is to demonstrate how the adversarial style transfer helps learn a robust policy and improves the robustness of the base RL policy used. 
Moreover, in our case, the final policy is still trained using original observation. Thus this policy can be regularized using the method proposed in DRAC \citep{raileanu2020automatic}. The general data augmentation can still be used on observation before passing it to policy. Thus our method might be applied to many existing algorithms, including data augmentation and regularization. Combining our method with these algorithms could improve performance; investigating them can be an interesting avenue for future work.

%% file: conclusion.tex
\section{Discussion}
In summary, we proposed an algorithm, Adversarial Robust Policy Optimization (ARPO), to improve generalization for reinforcement learning agents by removing the effect of overfitting in high-dimensional observation space. Our method consists of a max-min game theoretic objective where a generator is used to transfer the style of high-dimensional observation (e.g., image), thus perturb original observation while maximizing the agent's probability of taking a different action for the corresponding input observation. In contrast, a policy network updates its parameters to minimize the effect of such translation, thus being robust to the perturbation while maximizing the expected future reward. We tested our approaches on Procgen and distracting control benchmarks in generalization setup and confirmed the effectiveness of our proposed algorithm. Furthermore, empirical results show that our method generalizes better in unseen test environments than the tested baseline algorithms.

Note that the lack of visual diversity in the observation of the environment might result in poor clustering, which eventually leads to a less challenging style translation by the adversarial generator. 
However, in those scenarios, the reinforcement learning algorithms often perform well as the train and test environment remain consistent (e.g., same background). Furthermore, this setup might lead the agent to overfit the training environment, which results in the agent performing poorly in a slightly different environment \citep{Song2020Observational,zhang2018study}. In this paper, we are interested in the scenario where the train and test environment are visually different, which we think is a more practical setup \citep{cobbe2020leveraging,stone2021distracting}. 

%% file: appendix.tex
\section{Additional Procgen Results}
\label{sec:procgen_results}
\begin{figure*}[!ht]
\centering
\includegraphics[width=0.999\linewidth]{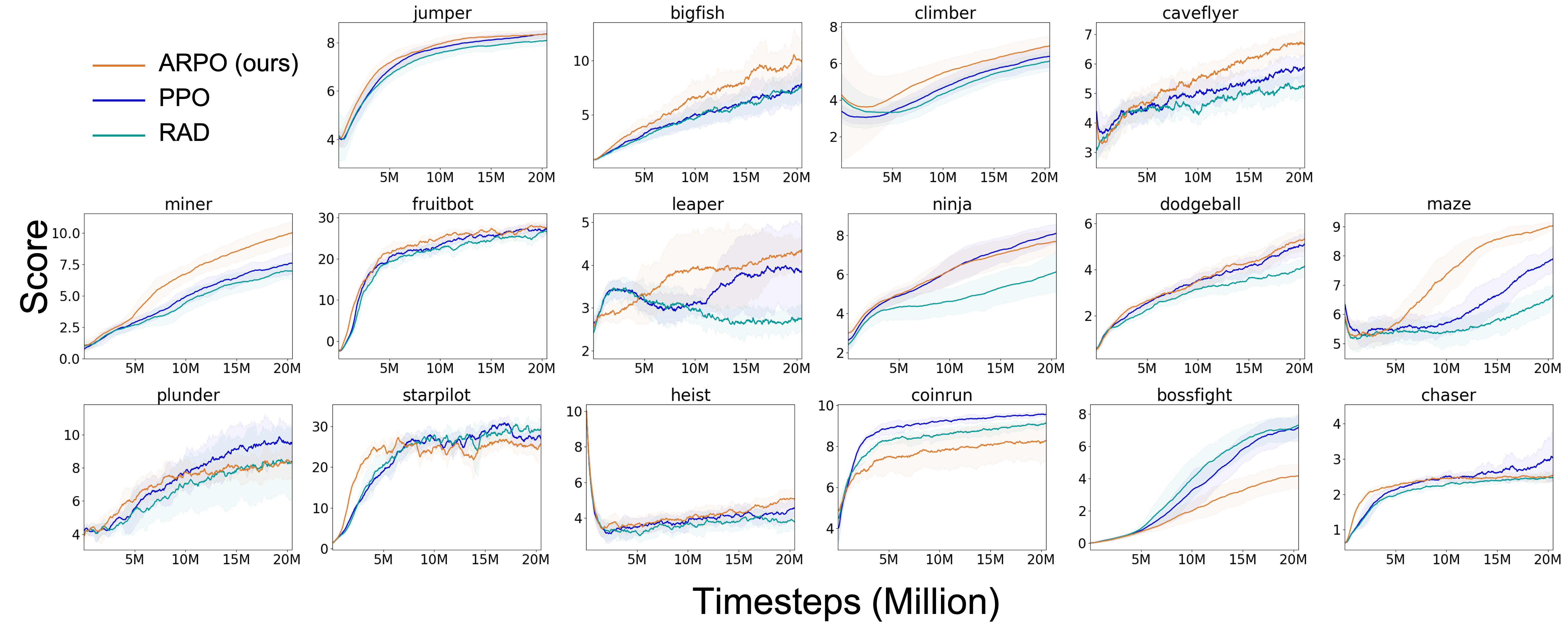} 
\caption{\small  
Sample efficiency results. Train performance (learning curve) comparison on Procgen environments. Despite using perturbation on the observation, our agent ARPO outperforms the baselines in many environments, thus achieving better sample efficiency during training. The results are averaged over 3 seeds.
}
\centering
\label{fig:arpo_procgen_train}
\end{figure*}

Figure \ref{fig:arpo_procgen_train} shows the sample efficiency results for ARPO, PPO, and RAD.

\section{Additional Distracting Control Results}
\label{sec:add_dist_control_results}
\begin{figure*}[!ht]
\centering
\includegraphics[width=0.99\linewidth]{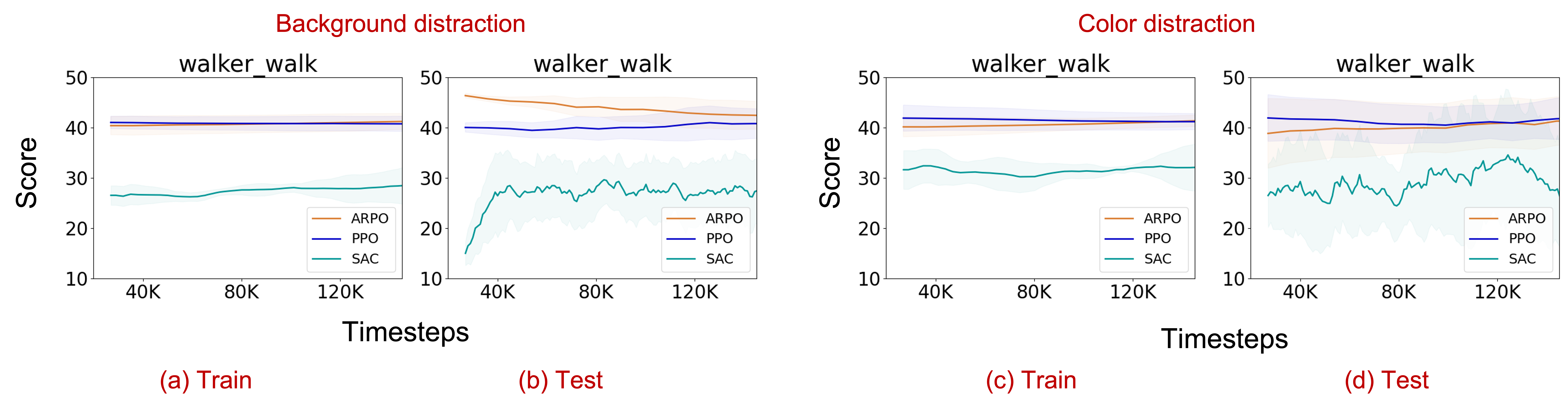} 
\caption{\small Distracting control Walker walk environment results (including SAC). The results are averaged over 3 seeds. \textbf{(a, c)} Sample efficiency (train) \textbf{(b, d)} Generalization (test).}
\label{fig:ddmc_walker_walk_sac}
\centering
\end{figure*}

\noindent\textit{Comparison with SAC on Walker Walk}.
Results comparison on Distracting Control for ARPO, PPO, and SAC are in Figure \ref{fig:ddmc_walker_walk_sac} (Walker walk).
We see that ARPO (and PPO) perform better compared to the SAC in both sample efficiency (train) and generalization (test) in these settings.

\noindent\textit{Results on Cheetah run with background distraction}.

\begin{figure*}[!ht]
\centering
\includegraphics[width=0.99\linewidth]{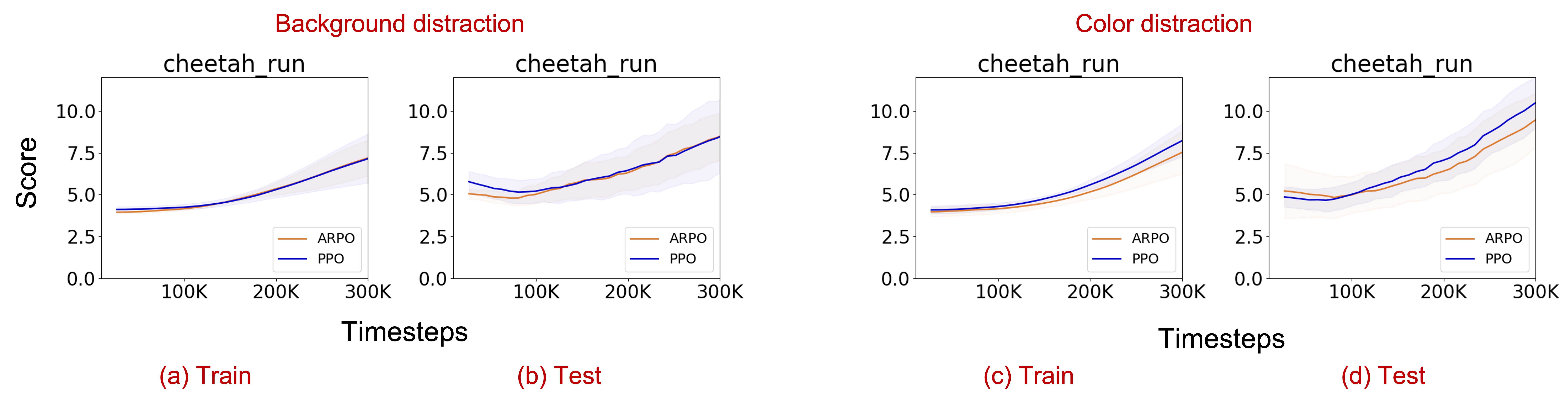} 
\caption{\small Distracting control Cheetah run environment results. The results are averaged over 3 seeds. \textbf{(a, c)} Sample efficiency (train), and \textbf{(b, d)} Generalization (test).}
\label{fig:ddmc_cheetah_run}
\centering

\end{figure*}

Results for ARPO and PPO on Cheetah run with background distraction are in Figure \ref{fig:ddmc_cheetah_run}. We see that both agents perform comparably in both background and color distraction. We observe similar results for both the agents in this setup of Cheetah run environments. Note that in this training timestep the PPO agents perform poorly in the training and testing. The highest achievable reward for this environment is much higher (see \cite{stone2021distracting}). Consequently the the ARPO agent could not improve over the base algorithm PPO. This suggest that the base algorithm needs to trained sufficiently to achieve any reasonable reward before we benefit of the adversarial training.

Note that the resulting score is low compared to the reported results in the benchmark paper \cite{stone2021distracting} which suggests the PPO itself could not learn helpful behavior in this environment.

\begin{figure*}[!ht]
\centering
\includegraphics[width=0.99\linewidth]{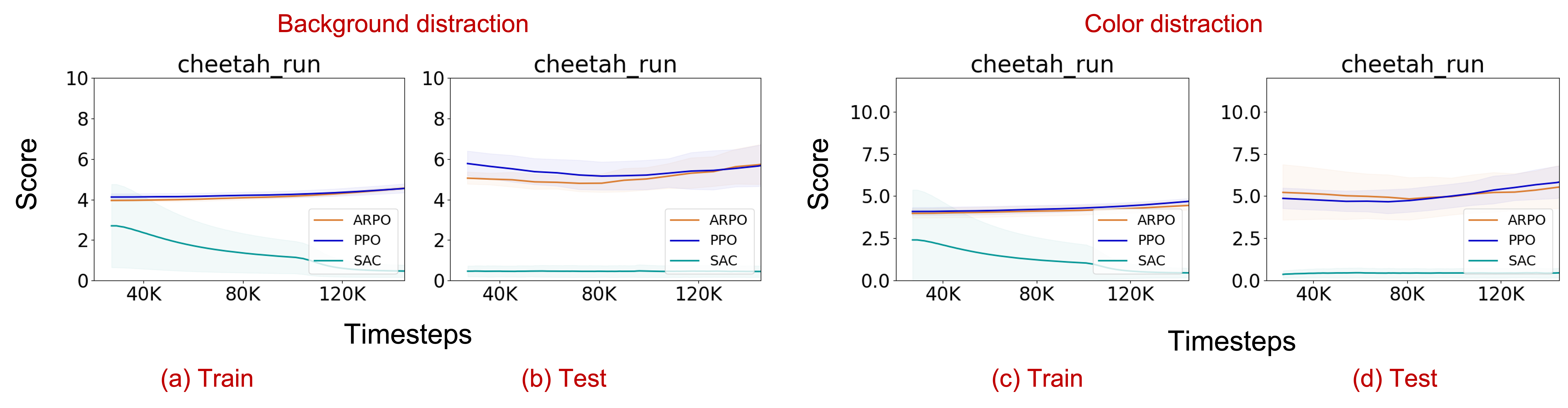} 
\caption{\small Distracting control Cheetah run environment results (including SAC). The results are averaged over 3 seeds. \textbf{(a, c)} Sample efficiency (train), and \textbf{(b, d)} Generalization (test).}
\label{fig:ddmc_cheetah_run_sac}
\centering
\end{figure*}

\noindent\textit{Comparison with SAC on Cheetah Run} 
\label{sec:sac_results_cheetah}
Results comparison on Distracting Control for ARPO, PPO, and SAC on Cheetah run are in Figure \ref{fig:ddmc_cheetah_run_sac}.

\section{Ablation Study and Hyperparameters}
\label{sec:ablation}

\subsection{Hyperparameters of ARPO}
We conduct a study on how ARPO agent's performance varies due to its hyperparameters: (a) the number of clusters generated by the GMM model for generator training, and (b) $\beta_1$, $\beta_2$ which indicate the amount of participation of policy and generator in the adversarial objective.

\begin{figure}
\begin{minipage}{\linewidth}
     \centering
     \begin{subfigure}[b]{0.49\textwidth}
         \centering
         \includegraphics[width=\textwidth]{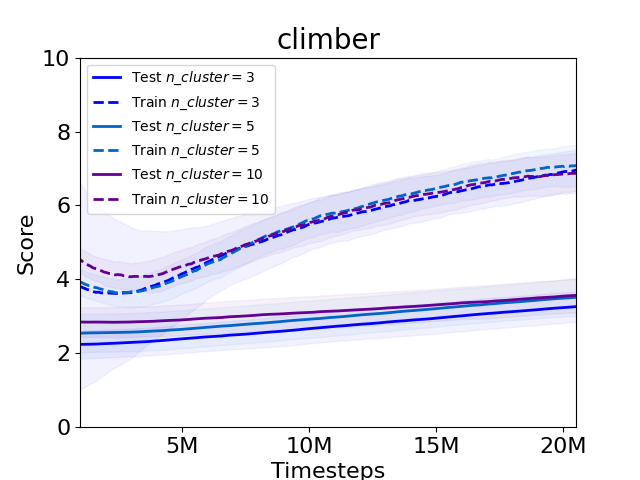}
         \caption{}
         \label{fig:procgen_arpo_ablation_ncluster_climber_200}
     \end{subfigure}
     \hfill
     \begin{subfigure}[b]{0.49\textwidth}
         \centering
         \includegraphics[width=\textwidth]{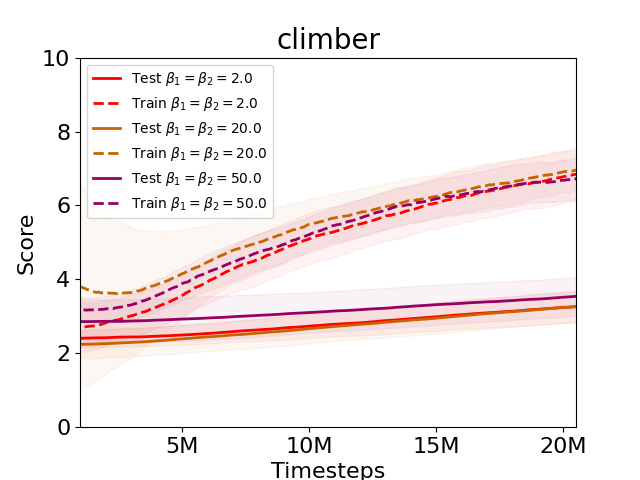}
         \caption{}
         \label{fig:procgen_arpo_ablation_beta_climber_200}
     \end{subfigure}
     \end{minipage}
        \caption{\small The results are averaged over 3 seeds. \textbf{(a)} Our ARPO agent's results on different cluster numbers. It improves generalization (test) performance with the increase in cluster number in the Climber environment. \textbf{(b)} Our ARPO agent's results on different $\beta$ values which determine participation on adversarial optimization.}
        \label{fig:arpo_procgen_ablation}
\end{figure}
Figure \ref{fig:arpo_procgen_ablation} shows the ablation results. We observe that our ARPO agent shows improvement in generalization (test) performance with the increase in number of cluster on Climber environment (Figure \ref{fig:procgen_arpo_ablation_ncluster_climber_200}). As the number of cluster increase, the generator has more option to choose for translation, and thus the policy might face a hard challenge and thus learn a more robust policy. In this paper, we report the ARPO agent's results on cluster $n\_cluster=3$.
On the other hand,  and ablation results on different $\beta$ values are shown in Figure \ref{fig:procgen_arpo_ablation_beta_climber_200}.
When $\beta$ values are large the test performance goes up; however, training performance suffer a bit. 
In this paper, we report the results on $\beta_1=\beta_2=20$, which shows a balance in train and test performance.

\subsection{KL Regularization}
\begin{figure*}[!ht]
\centering
\includegraphics[width=0.99\linewidth]{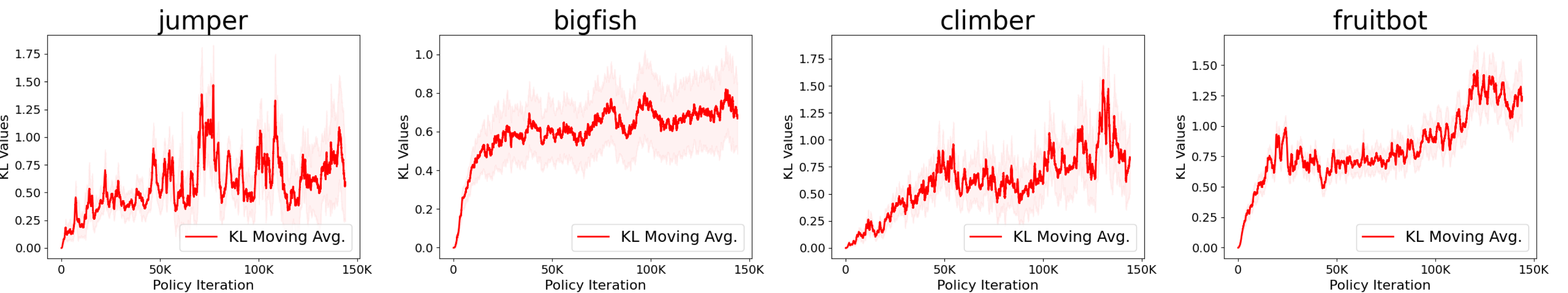} 
\caption{\small  
KL regularization during RL training for four Procgen environments. At the beginning of the training, both the policy and generator behave randomly or outputting same action for all the images; thus, the KL values become nearly zero. Gradually, the KL started to increase as the policy learning progressed with the adversarial generated observation. Eventually, the policy tries to adjust to the changes in the observation, and the KL values become stable (\textbf{bigfish}) or going downward (\textbf{jumper}, and \textbf{climber}). We see that in \textbf{fruitbot} environment, the values started to increase at the end, which suggests a possible divergence. This scenario might be an explanation of why the generalization performance started to drop at the end in Figure \ref{fig:arpo_procgen_test}, possible overfitting. Thus this KL measure might be a tool to detect possible divergence and overfitting. However, each environment's stability cutoff time (policy iteration) is different, suggesting that the KL regularization behavior might be different depending on the environment complexity. Ideally, with enough training, the KL values should converge to zero (again) if the policy reaches optimal (for true MDP state) and the generator fully recovers the true state from the observation and changes all the irrelevant information from the observation. Note that we limit the policy to train until a cutoff time (20 Million RL timesteps).
}
\centering
\label{fig:procgen_kl_results}
\end{figure*}

Figure \ref{fig:procgen_kl_results} shows the KL regularization results in different policy iteration on four Procgen environments. For each environment, we run the experiment and collect KL value ($KL(\pi_\theta(.|x_t), \pi_\theta(.|x_t'))$) for each policy iteration. To calculate moving average, we select a window size of 500 and the standard deviation is calculated on these 500 values, and showed in the shaded area in Figure  \ref{fig:procgen_kl_results}. Results analysis is in the caption.

\section{Additional Qualitative Results}
\begin{figure}[!ht]
\centering
\includegraphics[width=0.99\columnwidth]{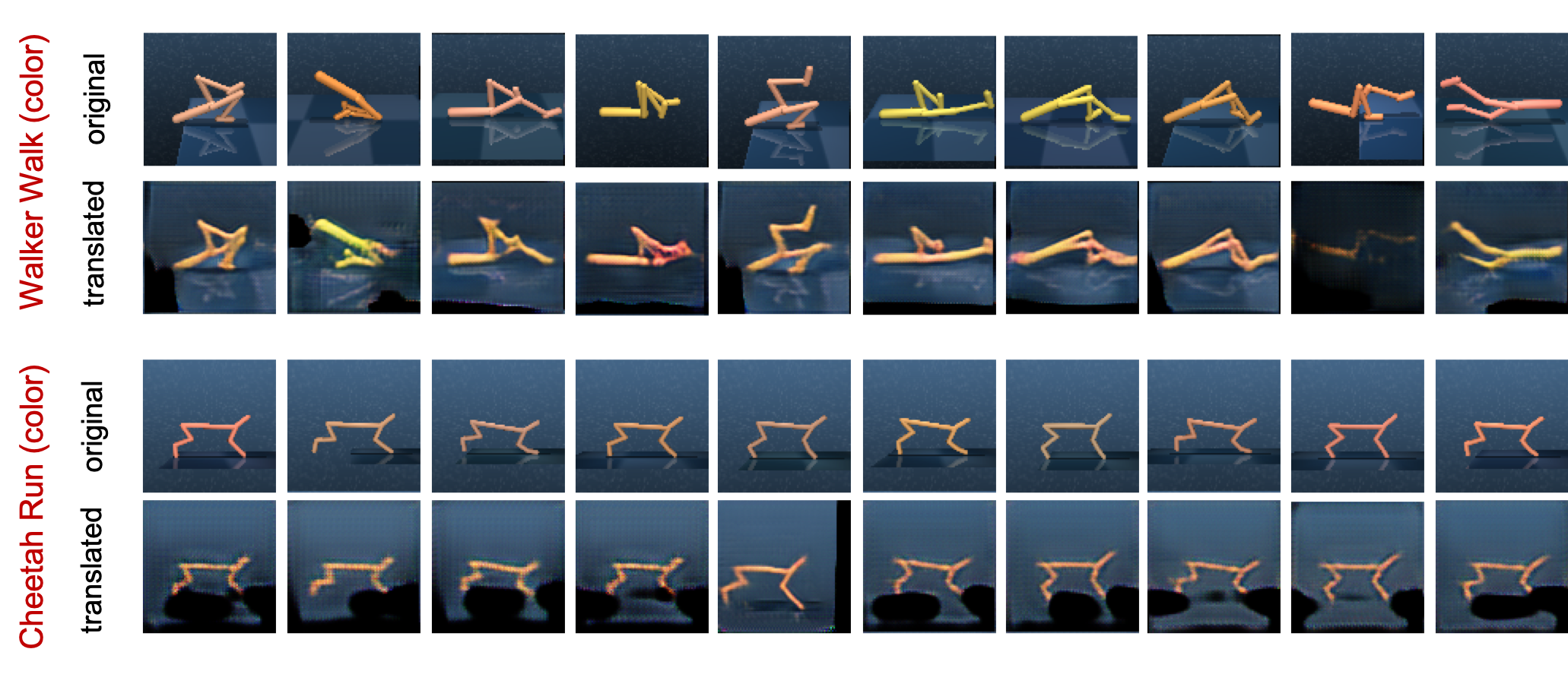} 
\caption{\small  
Sample translation of generator on \textbf{color} distraction for Walker-walk and Cheetah-run environments from Distracted Control Suite benchmark.
}
\centering
\label{fig:ddmc_color_translated_sample}
\end{figure}

Figure \ref{fig:ddmc_color_translated_sample} shows sample translation from the generator on color distracted environments.

\section{Training Loss of StarGAN Generator}

The adversarial loss is calculated as the Wasserstein GAN \citep{arjovsky2017wasserstein} objective with gradient penalty which is defined as 

\begin{equation} \label{eq:loss_disciminator}
\begin{split}
 \mathcal{L}_{adv} = \mathbb{E}_x[D_{src}] - \mathbb{E}_{x,c}[D_{src}(G(x,c))] \\
 - \lambda_{gp}  \mathbb{E}_{\hat{x}}[(||\nabla_{\hat{x}}D_{src}(\hat{x})|| - 1)^2],
\end{split}
\end{equation}

where $\hat{x}$ is sampled uniformly along a straight line between a pair of real and generated fake images. Here, the gradient penalty helps to stabilize the GAN training process and has been shown to perform better than traditional GAN training \citep{choi2018stargan,arjovsky2017wasserstein,gulrajani2017improved}. We set the hyperparameter $\lambda_{gp}=10$.

The classification loss of real image is defined as 
\begin{equation} \label{eq:loss_class_real}
    \mathcal{L}^r_{cls} = \mathbb{E}_{x, c'}[-\log D_{cls}(c'|x)]
    ,
\end{equation}
where $D_{cls}(c'|x)$ is the probability distribution over all clusteer labels.
Similarly, the classification loss of fake generated image is defined as 
\begin{equation} \label{eq:loss_class_real}
    \mathcal{L}^f_{cls} = \mathbb{E}_{x, c}[-\log D_{cls}(c|G(x, c))]
    ,
\end{equation}

The reconstruction loss with the generator objective defined as 
\begin{equation} \label{eq:loss_generator_rec}
    \mathcal{L}_{rec} = \mathbb{E}_{x, c, c'}[||x-G(G(x,c), c')||_1], 
\end{equation}
where we use the $L1$ norm.

\section{Environment Details}

\noindent\textbf{Procgen} 
We conducted experiments on OpenAI Procgen \cite{cobbe2020leveraging} benchmark, consisting of diverse procedurally-generated environments with different action sets. 
This environment has been used to measure how quickly (sample efficiency) a reinforcement learning agent learns generalizable skills. 
These environments greatly benefit from the use of procedural content generation, the algorithmic creation of a near-infinite supply of highly randomized content. 
The design principles consist of high diversity, tunable difficulty, shared action, shared observation space, and tunable dependence on exploration. 
Procedural generation logic directs the level layout and other game-specific details. Thus, to master any of these environments, agents must learn an effective policy across all environment variations. We use all 16 environments available in this benchmarks.

All environments use a discrete 15 dimensional action space which generates $64 \times 64 \times 3$  RGB image observations. Note that some environments may use no-op actions to accommodate a smaller subset of actions.

\section{Implementation Details}
\noindent\textbf{Procgen Experiments}
For experimenting on Procgen environments, we used RLlib \cite{liang2018rllib} to implement our ARPO, and baselines PPO and RAD cutour color algorithms. For all the agents' policy network (model), we use a CNN architecture used in IMPALA \cite{espeholt2018impala} which is the best performing model in Procgen benchmark \citep{cobbe2020leveraging}. We use the same policy parameters for all agents for a fair comparison.

Policy learning hyperparameter settings for all the agents (ARPO, PPO, and RAD) are set same for fair comparison and detect the effect of our proposed method (max-min adversarial objective with perturbation network). These hyperparameters are given in Table \ref{tab:rllib-hyp}. Note that only the custom parameters are given here, other defaults parameter values can be found in the RLlib library \cite{liang2018rllib}.

\begin{table}
\caption{Hyperparameters for Procgen (RLlib) Experiments} 
\label{tab:rllib-hyp} 
\begin{center}
 \scriptsize
\begin{tabular}{c|c}
 Description & Hyperparameters  \\
\hline
 Discount factor of the MDP & $gamma: 0.999$  \\
\hline
The GAE(lambda) parameter & $lambda: 0.95$  \\
\hline
The default learning rate & $lr: 5.0e-4$ \\
\hline
Number of epochs per train batch & $num\_sgd\_iter: 3$ \\
\hline
Total SGD batch size & $sgd\_minibatch\_size: 2048$ \\
\hline
Training batch size & $train\_batch\_size: 16384$ \\
\hline
Initial coefficient for KL divergence &  $kl\_coeff: 0.0$ \\
\hline
Target value for KL divergence & $kl\_target: 0.01$ \\
\hline
Coefficient of the value function loss & $vf\_loss\_coeff: 0.5$ \\
\hline
Coefficient of the entropy regularizer & $entropy\_coeff: 0.01$\\
\hline
PPO clip parameter & $clip\_param: 0.2$ \\
\hline
Clip param for the value function & $vf\_clip\_param: 0.2$ \\
\hline
Clip the global amount & $grad\_clip: 0.5$ \\
\hline
Default preprocessors & $deepmind$ \\
\hline
PyTorch Framework & $framework: torch$ \\
\hline
Settings for Model & $custom\_model:impala\_cnn\_torch$ \\
\hline
Rollout Fragment & $rollout\_fragment\_length: 256$ \\
\hline
\end{tabular}
\end{center}
\end{table}

\noindent\textbf{Distracting Control Experiments}
For experimenting on Distracting Control Suite \citep{stone2021distracting} environments, we used RLlib \cite{liang2018rllib} to implement our ARPO, and baselines PPO and SAC algorithms. We use the PPO's CNN-based policy network (model) from RLlib for ARPO, and PPO. The policy specific parameters for ARPO, and PPO are the same which are given in Table \ref{tab:rllib-arpo-ppo}.

For SAC, we use it's CNN-based model available in RLlib. The policy specific parameters are given in Table \ref{tab:rllib-sac}.
Note that only the custom parameters for the RLlib implementation are given here, other defaults parameter values can be found in the RLlib library \cite{liang2018rllib}.

\begin{table}
\caption{ARPO, and PPO Hyperparameters for Distracting Control Experiments}
\label{tab:rllib-arpo-ppo} 
\begin{center}
 \scriptsize
\begin{tabular}{c|c}
 Description & Hyperparameters  \\
\hline
Number of epochs per train batch & $num\_sgd\_iter: 3$ \\
\hline
Total SGD batch size & $sgd\_minibatch\_size: 256$ \\
\hline
Training batch size & $train\_batch\_size: 8192$ \\
\hline
PyTorch Framework & $framework: torch$ \\
\hline
\end{tabular}
\end{center}
\end{table}

\begin{table}
\caption{SAC Hyperparameters for Distracting Control Experiments}
\label{tab:rllib-sac} 
\begin{center}
 \scriptsize
\begin{tabular}{c|c}
Description & Hyperparameters  \\
\hline
Training batch size & $train\_batch\_size: 512$ \\
\hline
Timesteps per iteration & $timesteps\_per\_iteration: 1000$ \\
\hline
Timesteps per iteration & $learning\_starts: 5000$ \\
\hline
PyTorch Framework & $framework: torch$ \\
\hline
\end{tabular}
\end{center}
\end{table}

\noindent\textbf{Computing details}.
We used the following machine configuration to run our experiments: 20 core-CPU with 256 GB of RAM, CPU Model Name: Intel(R) Xeon(R) Silver 4114 CPU @ 2.20GHz, and a Nvidia A100 GPU.

\section{Comparison with DRAC}
\begin{figure}[!ht]
\centering
\includegraphics[width=0.990\columnwidth]{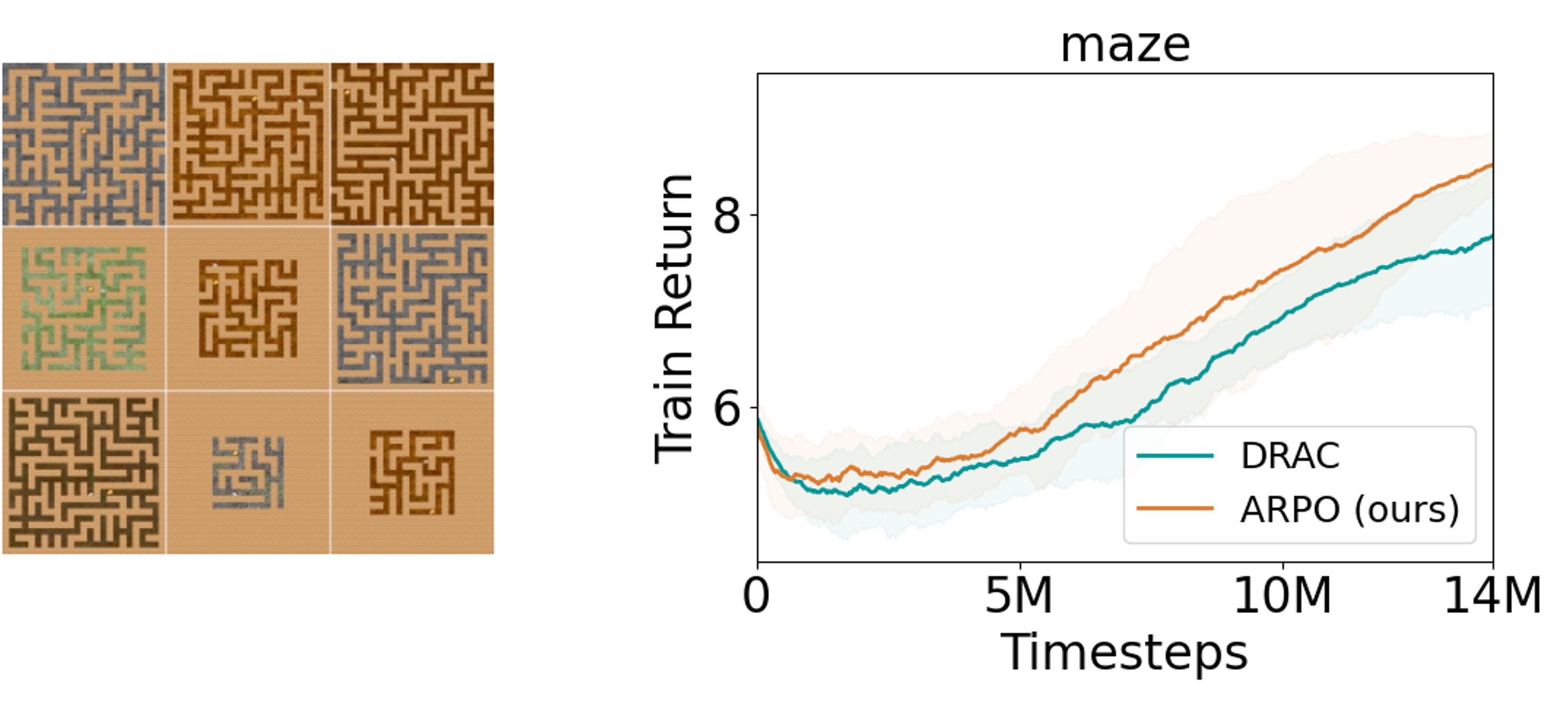} 
\caption{\small  
Training curve comparison with DRAC on Procgen Maze environment.
}
\centering
\label{fig:arpo_drac_train_maze}
\end{figure}

Figure \ref{fig:arpo_drac_train_maze} shows comparison with DRAC \cite{raileanu2020automatic} on Procgen Maze environment. We see that our method ARPO performs better in sample efficiency during training compared to DRAC. The results are averaged over 2 random seed runs.
We further show final results after training agents for 14M timesteps. Table \ref{tab:arpo_drac_maze} shows mean and standard deviation of results. For both the agents, the returns are smoothed using standard exponential averaging to account for the performance in previous timesteps.
\begin{table}
\caption{Results on Maze after 14M timesteps.} 
\label{tab:arpo_drac_maze} 
\begin{center}
\begin{tabular}{|c|c|c|c|c|}
 \hline
 \textbf{Eval} &\textbf{ ARPO (ours)} & \textbf{DRAC} \\
\hline
Train &  \textbf{8.55}$\pm$ \textbf{\small{0.29}} &  7.82$\pm$ \small{0.66} \\
\hline
Test &  \textbf{5.24} $\pm$ \textbf{\small{0.24}} &  5.22$\pm$ \small{1.38} \\
\hline
\end{tabular}
\end{center}
\end{table}
We see that our agent ARPO performs better in final results in train return compared to DRAC. On the other hand, both agents show similar mean values for test return. However, our method ARPO shows a smaller variance ($0.24$) compared to DRAC's high variance ($1.38$). These results demonstrate that overall our ARPO algorithm shows effective results compared to DRAC in the Procgen Maze environment. Note that in this setting DRAC uses tuned data augmentation (see the DRAC paper \cite{raileanu2020automatic}) which is \textit{crop} in case of maze. On the other hand, our method does not assume any such set of data augmentation. Instead, our method tries to figure out such perturbation on observation as the training progresses. 

\textit{Setup}. We implemented the DRAC method following the original paper \cite{raileanu2020automatic}. In particular, we incorporated policy and value function regularization following the implementation details in the original DRAC implementation. For a fair comparison, we keep all other hyperparameters the same as we did for all other experiments.

\section{ARPO on SAC}
In this experiment, we investigate the use of our technique on a different RL algorithm than PPO. We choose off-policy SAC to demonstrate the implication of our method. Following a similar setup as the PPO base algorithm, we regularize the policy using the KL term in Equation \ref{eq:policy}. The RL part and generator train together alternately, similar to the PPO base setup. We name this version of our agent \textbf{ARPO-SAC}.
\begin{figure}[!tbp]
  \centering
  \begin{minipage}[b]{0.49\textwidth}
    \includegraphics[width=0.99\textwidth]{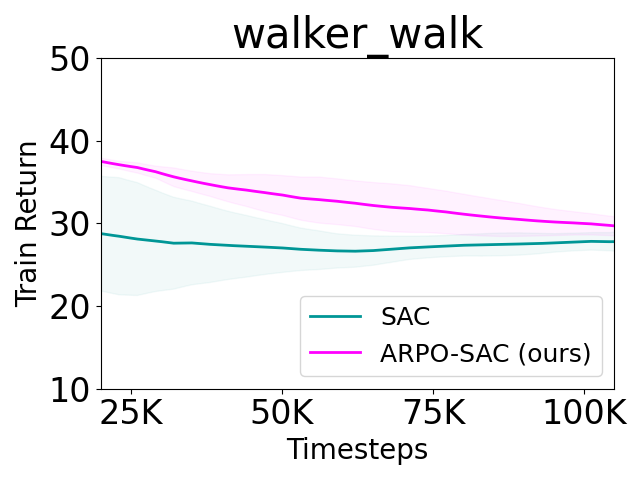}
  \end{minipage}
  \hfill
  \begin{minipage}[b]{0.49\textwidth}
    \includegraphics[width=0.99\textwidth]{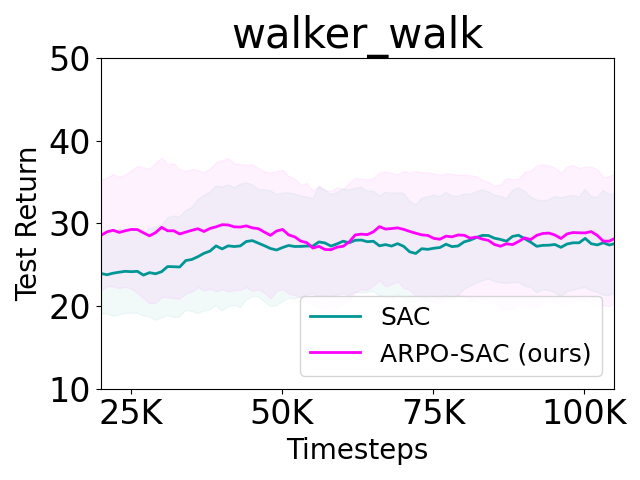}
  \end{minipage}
  \hfill
  \caption{\textbf{Results comparison of ARPO on SAC agent in Distracting control Walker walk environment with background distraction. Our agent ARPO with SAC base algorithm performs better in training in terms of train return. In test return, ARPO achieves slightly better generalization at the early timesteps and is similar in the rest of the timesteps. }
  }
  \label{fig:arpo_on_sac}
\end{figure}

Figure \ref{fig:arpo_on_sac} shows the results comparison on walker walk background distraction environment. Our agent ARPO-SAC shows better sample efficiency than SAC only baseline. It also shows slightly better generalization compared to SAC.

Note that PPO is an on-policy algorithm that directly learns the policy function. In contrast, SAC is an off-policy method that learns Q-functions and then generates policy from them. Thus, we see that policy gradient-based algorithms such as PPO better more amenable to being used with the adversarial setup. However, for the off-policy-based method such as SAC, Q-functions learning can be regularized with the KL-term, which might show an even better performance boost. However, investigating them in detail can be interesting future work.

\section{DCS Results with \textit{rliable} Metric}
We use the probabilistic algorithm comparison metric from rliable \cite{agarwal2021deep}.
\begin{figure}[!ht]
\centering
\includegraphics[width=0.990\columnwidth]{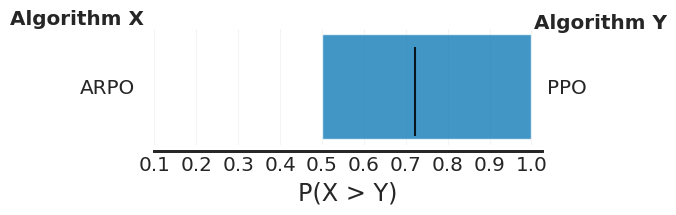} 
\caption{\small  
Probability comparison in walker-walk DCS for two detractors: Background video and color. We see that ARPO strongly outperforms PPO with a high probability, a chance of 50-100\%.
}
\centering
\label{fig:ddmc_comparison_probability}
\end{figure}
Figure \ref{fig:ddmc_comparison_probability} compares ARPO and PPO on Walker walk DCS environments. In particular, we use two environment setups: walker-walk color distraction and walker-walk background video distraction. The probability in Figure  \ref{fig:ddmc_comparison_probability} gives a range of probability values based on the performance of these algorithms in several random seeds. In this experiment, we use 3 random seed runs for each agent in each environment. Based on this metric, we see that ARPO strongly outperforms PPO with a high probability, a chance from 50-100\%. Overall, the results show ARPO improves performance over PPO in this setup.

\section{Visual diversity in training environments}
\begin{figure}[!ht]
\centering
\includegraphics[width=0.990\columnwidth]{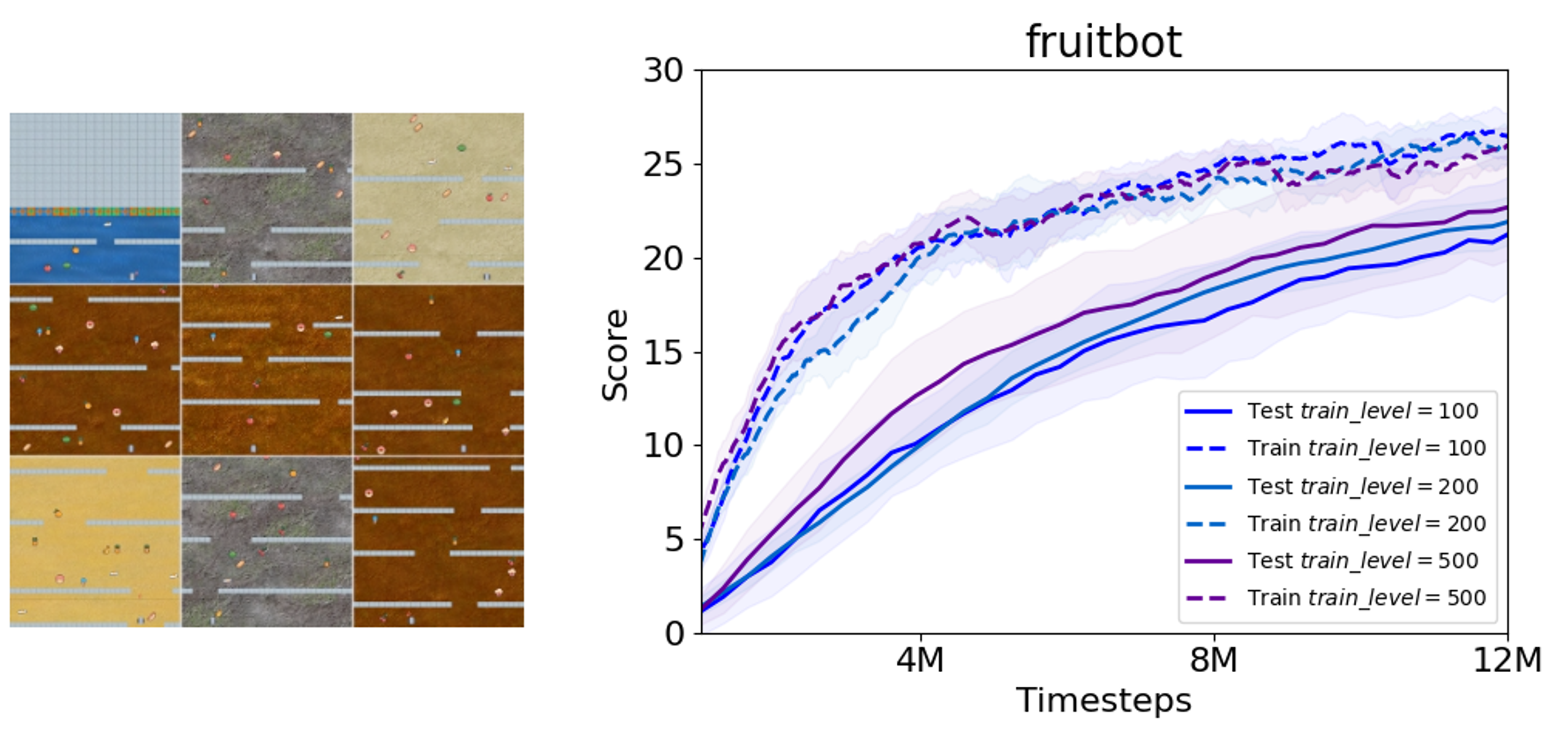} 
\caption{\small [\textbf{Left}] Some snippets of Procgen Fruitbot environmen. [\textbf{Right}] Performance comparison of various train level.
}
\centering
\label{fig:apro_train_diversity_fruitbot}
\end{figure}

We conducted experiments to investigate visual diversity in training environments. Figure \ref{fig:apro_train_diversity_fruitbot} shows the results on Procgen Fruitbot environment. We increase the training environment by allowing the agent to have more training levels. We choose Fruitbot due to its visual diversity (e.g., various background colors). For this setup, we choose different train levels 100, 200, and 500 during training and test on a fixed full distribution. We follow this use of full distribution as the test in the original Procgen benchmark papers \cite{cobbe2020leveraging} and many baselines evaluated on this benchmark \cite{raileanu2020automatic, laskin2020reinforcement}. Despite similar training performance, we observe that the test performance improved for 500 levels compared to 200 and 100.
Furthermore, the number of train levels 200 performs slightly better than 100 level in test score. However, the train results remain similar for all the agents. Note that one potential reason for improving the test performance (as observed in Procgen paper \cite{cobbe2020leveraging}). However, here we observe no such change in performance during training. Thus, the performance improvement might be due to diverge observation styles during training for our ARPO agent.

\section{Comparison with color jitter data augmentation}
\begin{figure}[!tbp]
  \centering
  \begin{minipage}[b]{0.49\textwidth}
    \includegraphics[width=0.99\textwidth]{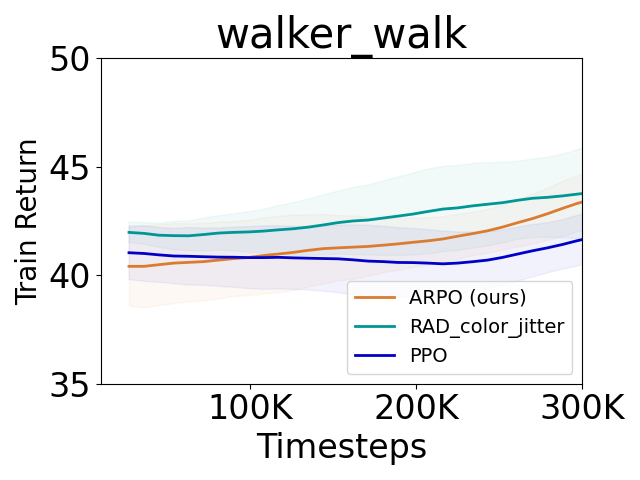}
  \end{minipage}
  \hfill
  \begin{minipage}[b]{0.49\textwidth}
    \includegraphics[width=0.99\textwidth]{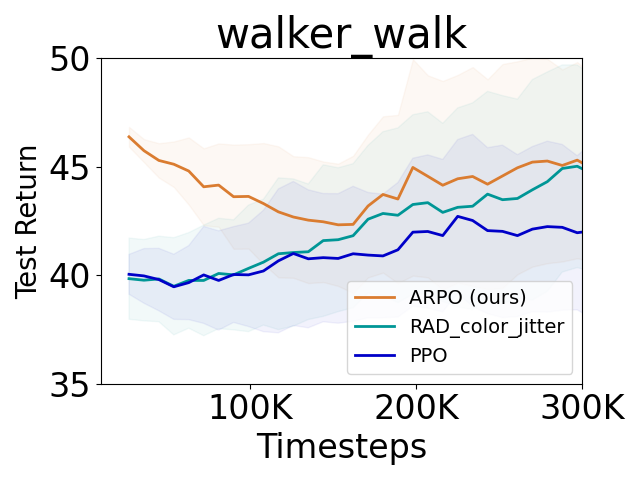}
  \end{minipage}
  \hfill
  \caption{\textbf{Comparison of our ARPO with RAD color jitter data augmentation. We see that color jitter augmentation performs slightly better in train return. However, in generalization (test return), our ARPO agent performs better, especially in the initial timesteps.}
  }
  \label{fig:arpo_vs_color_jitter}
\end{figure}

Figure \ref{fig:arpo_vs_color_jitter} shows results comparing our ARPO agent with color jitter data augmentation on a walker-walk background distraction environment. We follow RAD \cite{laskin2020reinforcement} approach to use the data augmentation. 
In particular, the observation is augmented using color jitter and then pass it to the agent. The results are averaged over 3 random seed runs. For a fair comparison, all the RL, environment, and hyperparameters are kept the same as PPO and ARPO.